\documentclass[aos,nameyear,dvips]{arximspdf}
\usepackage{subenv}
\usepackage{graphicx}

\doi{10.1214/09-AOS776}
\volume{39}
\issue{1}
\pubyear{2011}
\firstpage{1}
\lastpage{47}

\makeatletter
\newtheorem{theos}{Theorem}
\newtheorem{lems}{Lemma}

\newtheorem{cors}{Corollary}

\newproclaim{exas}{Example}
\newproclaim{remark}{Remark}
\newcommand{\bZ}{Z}
\newcommand{\bz}{z}
\newcommand{\bU}{U}
\newcommand{\bX}{X}
\newcommand{\bY}{Y}
\newcommand{\epsilonn}{\varepsilon}

\def\sfrac#1#2{#1/#2}

\def\sklfrac#1#2{(#1/#2)}
\newcommand{\EE}{\mathbb{E}}
\newcommand{\PP}{\mathbb{P}}
\newcommand{\underset}[2]{\mathop{#2}\limits_{#1}}
\newcommand{\indep}{\perp\!\!\!\perp}
\newcommand{\eqdis}{\stackrel{d}{=}}

\newcommand{\mutinc}{\gamma}
\newcommand{\myerror}{\rho}

\makeatother

\begin{document}
\begin{frontmatter}

\title{Support union recovery in high-dimensional multivariate
regression\thanksref{T1}}
\runtitle{Support union recovery in multivariate regression}
\thankstext{T1}{Supported in part by NSF Grants
DMS-06-05165 and CCF-05-45862 to MJW, and by NSF Grant 05-09559 and DARPA
IPTO Contract FA8750-05-2-0249 to MIJ.}

\begin{aug}
\author[a]{\fnms{Guillaume} \snm{Obozinski}\corref{}\ead[label=e1]{guillaume.obozinski@ens.fr}},
\author[a]{\fnms{Martin J.} \snm{Wainwright}\ead[label=e2]{wainwrig@stat.berkeley.edu}} \and\break
\author[a]{\fnms{Michael I.} \snm{Jordan}\ead[label=e3]{jordan@stat.berkeley.edu}}
\runauthor{G. Obozinski, M. J. Wainwright and M. I. Jordan}
\affiliation{INRIA---Willow Project-Team Laboratoire d'Informatique de
l'Ecole\\ Normale Sup\'{e}rieure,
%University of California at Berkeley,
University of California, Berkeley and\\
University of California, Berkeley}
\address[a]{G. Obozinski\\
M. J. Wainwright\\
M. I. Jordan\\
Department of Statistics\\
University of California, Berkeley\\
367 Evans Hall\\
Berkeley, California 94720\\
USA\\
\printead{e1}\\
\phantom{E-mail:\ }\printead*{e2}\\
\phantom{E-mail:\ }\printead*{e3}}
%adresu isvedimo komanda gale!
\end{aug}

% HISTORY:
\received{\smonth{6} \syear{2009}}
\revised{\smonth{11} \syear{2009}}

% ABSTRACT
%
\begin{abstract}
In multivariate regression, a $K$-dimensional response vector is
regressed upon a common set of $p$ covariates, with a matrix
$B^*\in\mathbb{R}^{p\times K}$ of regression coefficients.
We study the behavior of the \textit{multivariate group Lasso}, in which
block regularization
based on the $\ell_1/\ell_2$ norm is used for \textit{support union recovery},
or recovery of the set of $s$ rows for which $B^*$ is nonzero.
Under high-dimensional scaling, we show that the multivariate group
Lasso exhibits a
threshold for the recovery of the exact row pattern with high probability
over the random design and noise that is specified by the sample
complexity parameter $\theta(n, p, s)
:=
n/[2 \psi(B^*) \log(p- s)]$. Here $n
$ is
the sample size, and $\psi(B^*)$ is a \textit{sparsity-overlap
function} measuring a combination of the sparsities and overlaps of
the $K$-regression coefficient vectors that constitute the
model. We prove that the multivariate group Lasso succeeds for problem sequences
$(n, p, s)$ such that $\theta(n, p,
s)$ exceeds a critical level $\theta_u$, and fails for
sequences such that $\theta(n, p, s)$ lies below a
critical level $\theta_\ell$. For the special case of the standard
Gaussian ensemble, we show that $\theta_\ell= \theta_u$ so that the
characterization is sharp. The sparsity-overlap function
$\psi(B^*)$ reveals that, if the design is uncorrelated on the
active rows, $\ell_1/\ell_2$ regularization for multivariate regression
never harms performance relative to an ordinary Lasso approach and
can yield substantial improvements in sample complexity (up to a
factor of $K$) when the coefficient vectors are suitably orthogonal.
For more general designs, it is possible for the ordinary Lasso to
outperform the multivariate group Lasso. We complement our analysis with
simulations that demonstrate the sharpness of our theoretical results,
even for relatively small problems.
\end{abstract}

% KEYWORDS
%

\end{frontmatter}
%
%suskaldyti doi

%s1 ###
\section{Introduction}

The development of efficient algorithms for estimation of large-scale
models has been a major goal of statistical learning research in the
last decade. There is now a substantial body of work based on
$\ell_1$-regularization dating back to the seminal work
of~\citet{Tibshirani96} and Donoho and
collaborators~[\citet{Chen98}; \citet{Donoho01}]. The bulk
of this work has
focused on the standard problem of linear regression, in which one
makes observations of the form
%
%
%e1 ###
\begin{eqnarray}
\label{EqnLinReg}
y & = & \bX \beta^*+ w,
\end{eqnarray}
where $y \in\mathbb{R}^n$ is a real-valued vector of observations,
$w \in\mathbb{R}^n$ is an additive zero-mean noise vector and
$\bX
\in\mathbb{R}^{n\times p}$ is the design matrix. A subset of
the components of the unknown parameter vector $\beta^*\in
\mathbb{R}^p$ are assumed nonzero; the goal is to identify these
coefficients and (possibly) estimate their values. This goal can be
formulated in terms of the solution of the penalized optimization
problem
%
%e2 ###
\begin{equation}
\label{EqnL0}
\arg\min_{\beta\in\mathbb{R}^p}  \biggl\{ \frac{1}{n} \|y -
\bX \beta\|_2^2 + \lambda_n\|\beta\|_0  \biggr\},
\end{equation}
where $\|\beta\|_0$ counts the number of nonzero components in
$\beta$ and where $\lambda_n> 0$ is a regularization
parameter. Unfortunately, this optimization problem is
computationally intractable, a fact which has led various authors to
consider the convex relaxation~[\citet{Tibshirani96}; \citet{Chen98}]
%
%
%e3 ###
\begin{equation}
\label{EqnLasso}
\arg\min_{\beta\in\mathbb{R}^p}  \biggl\{ \frac{1}{n} \|y -
\bX \beta\|_2^2 + \lambda_n\|\beta\|_1  \biggr\},
\end{equation}
in which $\|\beta\|_0$ is replaced with the $\ell_1$ norm
$\|\beta\|_1$. This relaxation, often referred to as the
Lasso~[\citet{Tibshirani96}], is a quadratic program, and can be solved
efficiently by various methods~[e.g., \citet{Boyd02}; \citet{Osborne00}; \citet{lars}].

A variety of theoretical results are now in place for the Lasso, both
in the traditional setting where the sample size $n$ tends to
infinity with the problem size $p$ fixed~[\citet{Knight00}], as well
as under high-dimensional scaling, in which $p$ and $n$ tend
to infinity simultaneously, thereby allowing $p$ to be comparable
to or even larger than
$n$~[e.g., \citet{Meinshausen06}; \citet{Wainwright09}; \citet{MeiYu09}; \citet{BiRiTsy08}].
In many applications, it is natural to impose \textit{sparsity
constraints} on the regression vector $\beta^*$, and a variety of
such constraints have been considered. For example, one can consider
a ``hard sparsity'' model in which $\beta^*$ is assumed to contain
at most $s$ nonzero entries or a ``soft sparsity'' model in
which $\beta^*$ is assumed to belong to an $\ell_q$ ball with $q <
1$. Analyses also differ in terms of the loss functions that are
considered. For the model or variable selection problem, it is
natural to consider the zero--one loss associated with the
problem of recovering the unknown support set of $\beta^*$.
Alternatively, one can view the Lasso as a shrinkage estimator to be
compared to traditional least squares or ridge regression; in this
case, it is natural to study the $\ell_2$-loss $\|\widehat{\beta} -
\beta^*\|_2$ between the estimate $\widehat{\beta}$ and the ground
truth. In other settings, the prediction error $\mathbb{E}[(Y - X^T
\widehat{\beta})^2]$ may be of primary interest, and one tries to show
risk consistency (namely, that the estimated model predicts as well as
the best sparse model, whether or not the true model is sparse).

A number of alternatives to the Lasso have been explored in recent
years, and in some cases stronger theoretical results have been
obtained~[\citet{fan2001variable}; \citet{frank1993statistical}; \citet{huang2008asymptotic}].
However, the resulting optimization problems are generally nonconvex and
thus difficult to solve in practice. The Lasso remains a focus of
attention due to its combination of favorable statistical and
computational properties.

%s1.1 ###
\subsection{Block-structured regularization}

While the assumption of sparsity at the level of individual
coefficients is one way to give meaning to high-dimensional ($p
\gg n$) regression, there are other structural assumptions that
are natural in regression, and which may provide additional leverage.
For instance, in a hierarchical regression model, groups of regression
coefficients may be required to be zero or nonzero in a blockwise
manner; for example, one might wish to include a particular covariate
and all powers of that covariate as a group~[\citet{YuaLi06}; \citet{ZhaRocYu}].
Another example arises when we consider variable selection in the
setting of multivariate regression: multiple regressions can be
related by a (partially) shared sparsity pattern, such as when there
are an underlying set of covariates that are ``relevant'' across
regressions~[\citet{Obo07}; \citet{arg06}; \citet
{Turlach05}; \citet{Zhang08}]. Based on
such motivations, a recent line of
research~[\citet{Lan04}; \citet{Tropp06}; \citet
{YuaLi06}; \citet{ZhaRocYu}; \citet{Obo07}; \citet
{spamJRSS}] has studied
the use of \textit{block-regularization schemes}, in which the $\ell_1$
norm is composed with some other $\ell_q$ norm ($q > 1$), thereby
obtaining the $\ell_1/\ell_q$ norm defined as a sum of $\ell_q$ norms
over groups of regression coefficients. The best known examples of
such block norms are the $\ell_1/\ell_\infty$ norm~[\citet
{Turlach05}; \citet{Zhang08}] and the $\ell_1/\ell_2$
norm~[\citet{Obo07}].

In this paper, we investigate the use of $\ell_1/\ell_2$
block-regularization in the context of high-dimensional multivariate
linear regression, in which a collection of $K$ scalar outputs
are regressed on the same design matrix $\bX \in\mathbb
{R}^{n\times
p}$. Representing the regression coefficients as an $p\times
K$ matrix $B^*$, the multivariate regression model takes the
form
%
%
%e4 ###
\begin{eqnarray}
\label{EqnGenObsModel}
\bY& = & \bX  B^*+ W,
\end{eqnarray}
where $\bY\in\mathbb{R}^{n\times K}$ and $W\in
\mathbb{R}^{n\times K}$ are matrices of observations and
zero-mean noise, respectively. In addition, we assume a hard-sparsity
model for the regression coefficients in which column $k$ of the
coefficient matrix $B^*$ has nonzero entries on a subset
%
%
%e5 ###
\begin{equation}
S_k := \bigl\{ i \in\{1, \ldots,p\} \mid \beta^*_{ik} \neq0  \bigr\}
\end{equation}
of size $s_k :=|S_k|$.
In many applications it is natural to expect that the supports
$S_k$ should overlap. In that case, instead of estimating the
support of each regression separately, it might be beneficial to
first estimate the set of variables which are relevant to any
of the multivariate responses and to estimate only subsequently
the individual supports within that set. Thus we focus on the
problem of recovering the union of the supports, namely the set
$S:=\bigcup_{k=1}^KS_k$, corresponding to the
subset of indices $i \in\{1, \ldots, p\}$ that are involved
in at least one regression. We consider a range of problems in
which variables can be relevant to all, some, only one or
none of the regressions, and we investigate if and how the overlap
of the individual supports and the relatedness of individual
regressions benefit or hinder estimation of the \textit{support union}.

The \textit{support union problem} can be understood as the generalization
of the problem of variable selection to the group setting. Rather
than selecting specific components of a coefficient vector, we aim to
select specific rows of a coefficient matrix. We thus also refer to
the support union problem as the \textit{row selection problem}. We
note that recovering $S$, although not equivalent to recovering
each of the distinct individual supports $S_k$, addresses the essential
difficulty in recovering those supports. Indeed, as we show in
Section~\ref{sec:individual-supports}, given a method that returns
the row support $S$ with $|S| \ll p$ (with high probability),
it is straightforward to recover the individual supports $S_k$ by
ordinary least-squares and thresholding.

If computational complexity were not a concern, the natural way to
perform row selection for $B^*$ would be by solving the
optimization problem
%
%
%e6 ###
\begin{equation}
\label{EqnBlockRegProb0}
\arg\min_{B\in\mathbb{R}^{p\times K}}  \biggl\{
\frac{1}{2 n}  | \!  | \! | \bY- \bX B
| \!  | \!  |_{{F}}^2 + \lambda
_n
\| B \|_{\ell_{0} /\ell_{q}}  \biggr\},
\end{equation}
where $B =  ( \beta_{ik}  )_{1 \leq i \leq p, 1 \leq k
\leq K}$ is a $p\times K$ matrix, the quantity
$ | \!  | \! | \cdot  | \!  | \!  |_{{F}}$
denotes the Frobenius norm,\footnote{The
Frobenius norm of a matrix $A$ is given by $ | \!  | \! | A
 | \!  | \!  |_{{F}} :=
\sqrt{\sum_{i,j} A_{ij}^2}$.} and the ``norm''
$\| B \|_{\ell_{0} /\ell_{q}}$ counts the number of rows in $B$ that have
nonzero $\ell_q$ norm. As before, the $\ell_0$ component of this
regularizer yields a nonconvex and computationally intractable
problem, so that it is natural to consider the relaxation
%
%
%e7 ###
\begin{equation}
\label{EqnBlockRegProbq}
\arg\min_{B\in\mathbb{R}^{p\times K}}  \biggl\{
\frac{1}{2 n}  | \!  | \! | \bY-\bX B
| \!  | \!  |_{{F}}^2 + \lambda
_n
\| B \|_{\ell_{1} /\ell_{q}}  \biggr\},
\end{equation}
where $\| B \|_{\ell_{1} /\ell_{q}}$ is the block $\ell_1/\ell_q$ norm
%
%
%e8 ###
\begin{eqnarray}
\| B \|_{\ell_{1} /\ell_{q}} & :=& \sum_{i=1}^p
\Biggl(\sum_{j=1}^K{\beta_{ij}}^q \Biggr)^{\sfrac{1}{q}}=
\sum_{i=1}^p\|{\beta}_i\|_q.
\end{eqnarray}

The relaxation~(\ref{EqnBlockRegProbq}) is a natural generalization of
the Lasso; indeed, it specializes to the Lasso in the case $K=
1$. For later reference, we also note that setting $q = 1$ leads to
the use of the $\ell_1/\ell_1$ block norm in the
relaxation~(\ref{EqnBlockRegProbq}). Since this norm decouples across
both the rows and columns, this particular choice is equivalent to
solving $K$ separate Lasso problems, one for each column of the
$p\times K$ regression matrix $B^*$. A more
interesting choice is $q = 2$, which yields a block $\ell_1/\ell_2$ norm
that couples together the columns of $B$. Regularization with
the $\ell_1/\ell_2$ norm is commonly referred to as the \textit{group
Lasso} in the setting of univariate regression~[\citet{YuaLi06}]. We
thus refer to $\ell_1/\ell_2$ regularization in the multivariate setting
as the \textit{multivariate group Lasso}. Note that the multivariate group
Lasso can be viewed as a special case of the group Lasso, in that it
involves a specific grouping of regression coefficients, but the
multivariate setting brings new statistical issues to the fore.
As we discuss in Appendix~\ref{AppUniqueOpt}, the multivariate group
Lasso can be cast as a \textit{second-order cone program} (SOCP). This
is a family of convex optimization problems that can be solved
efficiently with interior point methods~[\citet{Boyd02}] and includes
quadratic programs as a particular case.

Some recent work has addressed certain statistical aspects of
block-regulariza\-tion schemes. \citet{Meier} perform an analysis
of risk consistency with block-norm regularization. \citet{Bach08}
provides an analysis of block-wise support recovery for the kernelized
group Lasso in the classical, fixed $p$ setting. In the
high-dimensional setting, \citet{spamJRSS} studies the consistency
of block-wise support recovery for the group Lasso for fixed design
matrices, and their result is generalized by \citet{Liu08} to
block-wise support recovery in the setting of general $\ell_1/\ell_q$
regularization, again for fixed design matrices. However, these
analyses do not discriminate between various values of $q$, yielding
the same qualitative results and the same convergence rates for $q=1$
as for $q>1$. Our focus, which is motivated by the empirical
observation that the group Lasso and the multivariate group Lasso can
outperform the ordinary Lasso~[\citet{Bach08}; \citet
{YuaLi06}; \citet{ZhaRocYu}; \citet{Obo07}],
is precisely the distinction between $q = 1$ and $q > 1$
(specifically $q = 2$). We note that in concurrent work~\citet{NegWai08} %, a subset of the current authors
have studied a related problem of support recovery
for the $\ell_1/\ell_\infty$ relaxation.

The distinction between $q = 1$ and $q = 2$ is also significant
from an optimiza\-tion-theoretic point of view. In particular, the SOCP
relaxations underlying the multivariate group Lasso ($q = 2$) are
generally tighter than the quadratic programming relaxation underlying
the Lasso ($q = 1$); however, the improved accuracy is generally
obtained at a higher computational cost~[\citet{Boyd02}]. Thus we can
view our problem as an instance of the general question of the
relationship of statistical efficiency to computational efficiency:
does the qualitatively greater amount of computational effort involved
in solving the multivariate group Lasso always yield greater
statistical efficiency? More specifically, can we give theoretical
conditions under which solving the generalized Lasso
problem~(\ref{EqnBlockRegProbq}) has greater statistical efficiency
than naive strategies based on the ordinary Lasso? Conversely, can
the multivariate group Lasso ever be worse than the ordinary Lasso?

With this motivation, this paper provides a detailed analysis of model
selection consistency of the multivariate group
Lasso~(\ref{EqnBlockRegProbq}) with $\ell_1/\ell_2$-regularization.
Statistical efficiency is defined in terms of the scaling of the
sample size $n$, as a function of the problem size $p$ and
of the
sparsity structure of the regression matrix $B^*$, required for
consistent row selection. Our analysis is high-dimensional in nature,
allowing both $n$ and $p$ to diverge, and yielding explicit
error bounds as a function of $p$. As detailed below, our
analysis provides affirmative answers to both of the questions above.
First, we demonstrate that under certain structural assumptions on the
design and regression matrix $B^*$, the multivariate group Lasso
is always guaranteed to outperform the ordinary Lasso, in that it
correctly performs row selection for sample sizes for which the Lasso
fails with high probability. Second, we also exhibit some problems
(though arguably not generic) for which the multivariate group Lasso
will be outperformed by the naive strategy of applying the Lasso
separately to each of the $K$ columns, and taking the union of
supports.

%s1.2 ###
\subsection{Our results}

The main contribution of this paper is to show that under certain
technical conditions on the design and noise matrices, the model
selection performance of block-regularized $\ell_1/\ell_2$
regression~(\ref{EqnBlockRegProbq}) is governed by the \textit{sample
complexity function}
%
%
%e9 ###
\begin{eqnarray}
\label{EqnOurContPar}
\theta_{\ell_1/\ell_2}(n, p ; B^*) & :=& \frac{n}{2 \psi(B^*)
\log(p-s)},
\end{eqnarray}
where $n$ is the sample size, $p$ is the ambient dimension,
$s= |S|$ is the number of rows that are nonzero and
$\psi(\cdot)$ is a \textit{sparsity-overlap function}. Our use of
the term ``sample complexity'' for $\theta_{\ell_1/\ell_2}$ reflects
the role it
plays in our analysis as the rate at which the sample size must grow
in order to obtain consistent row selection as a function of the
problem parameters. More precisely, for scalings $(n, p,
s, B^*)$ such that $\theta_{\ell_1/\ell_2}(n, p ; B^*)$
exceeds a fixed critical threshold $\theta_u \in(0, +\infty)$, we
show that the probability of correct row selection by the
$\ell_1/\ell_2$ multivariate group Lasso converges to one, and
conversely, for scalings such that $\theta_{\ell_1/\ell_2}(n, p ;
B^*)$ is below another threshold $\theta_\ell$, we show that the
multivariate group Lasso fails with high probability.

Whereas the ratio $(\log p)/n$ is standard for the
high-dimensional theory of $\ell_1$-regularization, the function
$\psi(B^*)$ is a novel and interesting quantity, one which
measures both the sparsity of the matrix $B^*$ as well as the
overlap between the different regressions, represented by the columns
of $B^*$ [see equation~(\ref{EqnDefnGenKeyFunc}) for the precise
definition of $\psi(B^*)$]. As a particular illustration,
consider the special case of a univariate regression with $K=
1$, in which the convex program~(\ref{EqnBlockRegProbq}) reduces to
the ordinary Lasso~(\ref{EqnLasso}). In this case, if the design
matrix is drawn from the standard Gaussian ensemble [i.e., $X_{ij}
\sim N(0,1)$, i.i.d.], we show that the sparsity-overlap function
reduces to $\psi(B^*) = s$, corresponding to the support
size of the single coefficient vector. We thus recover as a corollary
a previously known result~[\citet{Wainwright09}]: namely, the Lasso
succeeds in performing exact support recovery once the ratio
$n/[s\log(p-s)]$ exceeds a certain critical
threshold. At the other extreme, for a genuinely multivariate problem
with $K> 1$ and $s$ nonzero rows, again for a standard
Gaussian design, when the regression matrix is ``suitably
orthonormal'' relative to the design (see Section~\ref{SecMain} for a
precise definition), the sparsity-overlap function is given by
$\psi(B^*) = s/K$. In this case, $\ell_1/\ell_2$
block-regularization has sample complexity lower by a factor of
$K$ relative to the naive approach of solving $K$ separate
Lasso problems. Of course, there is also a range of behavior between
these two extremes, in which the gain in sample complexity varies
smoothly as a function of the sparsity-overlap $\psi(B^*)$ in the
interval $[\frac{s}{K}, s]$. On the other hand,
we also show that for suitably correlated designs, it is possible that
the sample complexity $\psi(B^*)$ associated with $\ell_1/\ell_2$
block-regularization is larger than that of the ordinary Lasso
($\ell_1/\ell_1$) approach.

The remainder of the paper is organized as follows. In
Section~\ref{SecMain}, we provide a precise statement of our main
results (Theorems~\ref{ThmMain} and~\ref{ThmNecessary}),
discuss some of their consequences and illustrate the close agreement
between our theoretical results and simulations.
Sections~\ref{SecProof} and~\ref{SecProofThmNecessary} are devoted to
the proofs of Theorems~\ref{ThmMain} and~\ref{ThmNecessary},
respectively, with the arguments broken down into a series of steps.
More technical results are deferred to the appendices. We conclude with
a brief discussion in Section~\ref{SecDiscuss}.

%s1.3 ###
\subsection{Notation}
We collect here some notation used throughout the paper.
For a (possibly random)
matrix $M\in\mathbb{R}^{p\times K}$, we define the Frobenius
norm\break $ | \!  | \! | M  | \!  | \!  |_{{F}} :
=(\sum_{i,j} m_{ij}^2)^{1/2}$, and for
parameters $1 \leq a \leq b \leq\infty$, the $\ell_a/\ell_b$ block norm
is defined as follows:
%
%
%e10 ###
\begin{eqnarray}
\| M \|_{\ell_{a} /\ell_{b}} & :=& \Biggl\{ \sum_{i=1}^p
\Biggl(\sum_{k=1}^K|m_{ik}|^b \Biggr)^{\sfrac{a}{b}}
\Biggr\}^{\sfrac{1}{a}}.
\end{eqnarray}
These vector norms on matrices should be distinguished
from the $(a,b)$-operator norms
%
%
%e11 ###
\begin{eqnarray}
|\!|\!| M | \! | \!|_{{a}, {b}} & :=& \sup_{\|x\|_b=1} \| Mx\|_a
\end{eqnarray}
(although some norms belong to both families; see
Lemma~\ref{LemGuillaume} in Appendix~\ref{AppLemGuillaume}).
Important special cases of the latter include the spectral norm
$|\!|\!| M | \! | \!|_{{2}, {2}}$ (also denoted $ | \! | \!
 | M  | \!  | \!  |_{{2}}$), and
the $\ell_\infty$-operator norm $|\!|\!| M | \! | \!|_{{\infty},
{\infty }} =
\max_{i=1, \ldots, p} \sum_{j=1}^K|M_{ij}|$, denoted
$ | \! | \! | M  | \!  | \!  |_{{\infty}}$ for short.

In addition to the usual Landau notation $\mathcal{O}$ and $o$, we write
$a_n=\Omega(b_n)$ for sequences such that $\frac{b_n}{a_n}=o(1)$.
We\vspace*{1pt} also use the notation $a_n=\Theta(b_n)$ if both $a_n=\mathcal{O}(b_n)$
and $b_n=\mathcal{O}(a_n)$ hold.

%%%%%%%%%%%%%%%%%%%%%%%%%%%%%%%%%%%%%%%%%%%%%%%%%%%%%%%%%%%%%%%%%%%%%%%

%s2 ###
\section{Main results and some consequences}
\label{SecMain}
The analysis of this paper considers the multivariate group Lasso
estimator, obtained
as a solution to the SOCP
%
%
%e12 ###
\begin{equation}
\label{EqnBlockRegProb}
\arg\min_{B\in\mathbb{R}^{p\times K}} \biggl\{
\frac{1}{2 n}  | \!  | \! | \bY-\bX B
| \!  | \!  |_{{F}}^2 + \lambda
_n
\| B \|_{\ell_{1} /\ell_{2}} \biggr\}
\end{equation}
for random ensembles of multivariate linear regression problems,
each of the form~(\ref{EqnGenObsModel}), where the noise matrix
$W\in\mathbb{R}^{n\times K}$ is assumed to consist
of i.i.d.\ elements $W_{ij} \sim N(0, \sigma^2)$. We consider
random design matrices $\bX $ with each row drawn in an
i.i.d.\ manner from a zero-mean Gaussian $N(0, \Sigma)$, where
$\Sigma\succ0$ is a $p\times p$ covariance matrix.
%We note in passing that analogs of our results hold for any design with
%sub-Gaussian rows; only the constants differ.\footnote{See~
%for an extended discussion of sub-Gaussian random vectors.}
Although the block-regularized problem~(\ref{EqnBlockRegProb}) need
not have a
unique solution in general, a consequence of our analysis is that in
the regime of interest, the solution is unique, so that we may talk
unambiguously about the estimated support $\widehat{S}$. The main
object of study in this paper is the probability $\mathbb{P}[\widehat{S}
= S]$, where the probability is taken both over the random choice
of noise matrix $W$ and random design matrix $\bX $. We study the
behavior of this probability as elements of the triplet $(n,
p, s)$ tend to infinity.

%s2.1 ###
\subsection{Notation and assumptions}

More precisely, our main result applies to sequences of models indexed
by $(n, p(n), s(n))$, an associated
sequence of $p\times p$ covariance matrices and a sequence
$\{B^*\}$ of coefficient matrices with row support
%
%
%e13 ###
\begin{eqnarray}
S& :=& \{ i \mid {\beta}^*_i \neq0
\}
\end{eqnarray}
of size $|S| = s=s(n)$. We use ${S^c}$ to
denote its complement (i.e., ${S^c}:=\{1, \ldots, p\}
\backslash S$). We let
%
%
%e14 ###
\begin{equation}
\label{EqnDefnBetamin}
b^*_{\min}:=\min_{i \in S} \|{\beta}^*_i\|_2
\end{equation}
correspond to the minimal $\ell_2$ row-norm of the coefficient matrix
$B^*$ over its non-zero rows. Given an observed pair $(Y, X)$
from the model~(\ref{EqnGenObsModel}), the goal is to estimate
the row support $S$ of the matrix $B^*$.

We impose the following conditions on the covariance $\Sigma$ of the
design matrix:
\begin{enumerate}[(A1)]
\item[(A1)] \textit{Bounded eigenspectrum}: There exist fixed constants
$C_{\min}> 0$ and\break$C_{\max}< +\infty$ such that all eigenvalues of the
$s\times s$ matrix $\Sigma_{SS}$ are
contained in the interval $[C_{\min}, C_{\max}]$.
\item[(A2)] \textit{Irrepresentable condition}: There exists a fixed
parameter $\mutinc\in(0,1]$ such that
\begin{eqnarray*}
 | \! | \! | \Sigma_{{S^c}S} (\Sigma_{S S})^{-1}  |
\!  | \!  |_{{\infty}} & \leq& 1 - \mutinc.
\end{eqnarray*}
\item[(A3)] \textit{Self-incoherence}: $ | \! | \! | (\Sigma_{S
S})^{-1}  | \!  | \!  |_{{\infty}}
\leq D_{\max}$ for some $D_{\max}< +\infty$.
\end{enumerate}

The lower bound involving $C_{\min}$ in assumption~(A1) prevents
excess dependence among elements of the design matrix associated with
the support $S$; conditions of this form are required for model
selection consistency or $\ell_2$ consistency of the Lasso. The upper
bound involving $C_{\max}$ in assumption~(A1) is not needed for
proving success but only failure of the multivariate group Lasso. The
%mutual
%incoherence assumption
irrepresentable condition
and self-incoherence assumptions are also well known from
previous work on variable selection consistency of the
Lasso~[\citet{Meinshausen06}; \citet{Tropp06}; \citet
{Zhao06}]. Although such
%incoherence
%irrepresentable
assumptions
are not needed in analyzing $\ell_2$ or risk consistency,
they are known to be necessary for variable selection consistency of
the Lasso. Indeed, in the absence of such conditions, it is always
possible to make the Lasso fail, even with an arbitrarily large sample
size. [See, however,~\citet{MeiYu09} for methods that weaken the
%incoherence
irrepresentable
condition.] Note that these assumptions are trivially
satisfied by the standard Gaussian ensemble $\Sigma= I_{p\times
p}$, with $C_{\min}= C_{\max}= 1$, $D_{\max}= 1$ and $\mutinc= 1$.
More generally, it can be shown that various matrix classes (e.g.,
Toeplitz matrices, tree-structured covariance matrices, bounded
off-diagonal matrices) satisfy these
conditions~[\citet{Meinshausen06}; \citet{Zhao06};
\citet{Wainwright09}].

We require a few pieces of notation before stating the main results.
For an arbitrary matrix $B_S \in\mathbb{R}^{s\times K}$
with $i$th row ${\beta}_i \in\mathbb{R}^{1 \times K}$, we
define the matrix \mbox{$\zeta(B_S) \in\mathbb{R}^{s\times K}$} with $i$th row
%
%
%e15 ###
\begin{eqnarray}
\label{EqnDefnZ}
\zeta({\beta}_i) & :=& \frac{{\beta}_i}{\|{\beta}_i\|_2},
\end{eqnarray}
%
% MJ we need to allow \Brow_i = 0 in the notation, no?
when ${\beta}_i \neq0$, and we set $\zeta({\beta}_i) = 0$ otherwise.
With this notation, the \textit{sparsity-overlap function} is given by
%
%
%e16 ###
\begin{eqnarray}
\label{EqnDefnGenKeyFunc}
\psi(B) & :=&  | \!  | \! | \zeta(B_S)^T (\Sigma
_{SS})^{-1} \zeta(B_S)  | \!  | \!  |_{{2}},
\end{eqnarray}
where $ | \!  | \! | \cdot  | \!  | \!
|_{{2}}$ denotes the spectral norm. We use this
sparsity-overlap function to define the \textit{sample complexity
parameter}, which captures the effective sample size
%
%
%e17 ###
\begin{eqnarray}
\label{EqnDefnContPar}
\theta_{\ell_1/\ell_2}(n, p; B^*) & :=& \frac{n}{2
\psi(B^*) \log(p-s)}.
\end{eqnarray}

In the following two theorems, we consider a random design matrix
$\bX $
drawn with i.i.d.\ $N(0, \Sigma)$ row vectors, where $\Sigma$
satisfies assumptions~(A1)--(A3), and an observation
matrix $\bY$ specified by model~(\ref{EqnGenObsModel}). In
order to
capture dependence induced by the design covariance matrix, for any positive
semidefinite matrix \mbox{$Q \succeq0$}, we define the quantities
%
%e18 ###
%
%
\setcounter{equation}{0}
\renewcommand{\theequation}{18\alph{equation}}
\begin{eqnarray}
\label{EqnRhoLower}
\rho_{\ell}(Q) & :=& \tfrac{1}{2} \min_{i \neq j} [Q_{ii} +
Q_{jj} - 2 Q_{ij}]\quad
\end{eqnarray}
and
\begin{eqnarray}
\label{EqnRhoUpper}
\rho_{u }(Q) & :=& \max_{i} Q_{ii}.
\end{eqnarray}
We note that by definition, we have $\rho_{\ell}(Q) \leq\rho_{u }(Q)$
whenever $Q \succeq0$. Our bounds are stated in terms of these
quantities as applied to the conditional covariance matrix
\begin{eqnarray*}
\Sigma_{S^c S^c | S} & :=& \Sigma_{{S^c}
{S^c}}-\Sigma_{{S^c}S} ({\Sigma}_{SS})^{-1} \Sigma_{S{S^c}}.
\end{eqnarray*}
\setcounter{equation}{18}
\renewcommand{\theequation}{\arabic{equation}}

Our first result is an achievability result, showing that the
multivariate group Lasso succeeds in recovering the row support and
yields consistency in $\ell_\infty/\ell_2$ norm. We state this result\vadjust{\goodbreak}
for sequences of regularization parameters
$\lambda_n=\sqrt{\frac{f(p) \log p}{n}}$, where $f(p)\underset{p\rightarrow+\infty}{\longrightarrow} +\infty$ is any
function such that $\lambda_n\rightarrow0$. We also assume
that $n$ is sufficiently large such that $s/n<
1/2$.
%t1
\begin{theos}
\label{ThmMain}
Suppose that we solve the multivariate group Lasso with specified
regularization parameter sequence $\lambda_n$ for a sequence of
problems indexed by $(n, p, B^*,\Sigma)$ that satisfy
assumptions \textup{(A1)--(A3)}, and such that,
for some $\nu>0$,
\begin{eqnarray}
\label{EqnUpperBound}
\theta_{\ell_1/\ell_2}(n, p; B^*)
= \frac{n}{2\psi(B^*) \log(p-s)}
> (1+\nu)\frac{\rho_{u }(\Sigma_{S^c S^c | S})}{\mutinc^2}.
\end{eqnarray}
Then for universal constants $c_i > 0$ (i.e., independent of
$n, p, s, B^*, \Sigma$), with probability
greater than $1 - c_2 \exp(-c_3 K\log s) - c_0 \exp(-c_1
\log(p-s))$, the following statements hold:
\begin{enumerate}[(a)]
\item[(a)] The multivariate group Lasso has a unique solution
$\widehat{B}$ with row support $S(B)$ that is contained
within the true row support $S(B^*)$, and moreover satisfies
the bound
%
%e19 ###
\begin{eqnarray}\label{EqnMyerror}
\| \widehat{B}-B^* \|_{\ell_{\infty} /\ell_{2}} & \leq&
\mathop{\underbrace{\sqrt{\frac{8 K\log s}{C_{\min}n}} +\lambda_nD_{\max}+ \frac{6 \lambda_n}{C_{\min}}\sqrt{\frac{s^2}{n}}_{}}}_{\myerror(n, s, \lambda_n)}.
\end{eqnarray}
\item[(b)] If $\frac{\myerror}{b^*_{\min}} = o(1)$, the estimate of the
row support, $S(\widehat{B}) :=\{ i \in\{1,
\ldots,
p\} \mid \widehat{\beta}_i \neq0 \}$, specified by this
unique solution is equal to the row support set $S(B^*)$ of
the true model.
\end{enumerate}
\end{theos}

Note that the theorem is naturally separated into two distinct but
related claims. Part (a) guarantees that the method produces \textit{no
false inclusions} and, moreover, bounds the maximum $\ell_2$-error
across the rows. Part (b) requires some additional
assumptions---namely, the restriction $\frac{\myerror}{b^*_{\min}} =
o(1)$ ensuring that the error $\myerror$ is of lower order than the
minimum $\ell_2$-norm $b^*_{\min}$ across rows---but also guarantees the
stronger result of no false exclusions as well, so that the method
recovers the row support exactly. Note that the probability of these
events converges to one only if both $(p- s)$ and
$s$ tend to infinity, which might seem counter-intuitive
initially (since problems with larger support sets $s$ might
seem harder). However, as we discuss at the end of
Section~\ref{SecUevent}, this dependence can be removed at the expense
of a slightly slower convergence rate for
$\| \widehat{B}-B^* \|_{\ell_{\infty} /\ell_{2}}$.

Our second main theorem is a negative result, showing that the
multivariate group Lasso fails with high probability if the rescaled
sample size $\theta_{\ell_1/\ell_2}$ is below a critical threshold.
In order to
clarify the phrasing of this result, note that Theorem~\ref{ThmMain}
can be summarized succinctly as guaranteeing that there is a unique
solution $\widehat{B}$ with the correct row support that satisfies
$\| \widehat{B} - B^* \|_{\ell_{\infty} /\ell_{2}} = o ( b^*_{\min
})$. The
following result shows that such a guarantee cannot hold if the sample
size $n$ scales too slowly relative to $p$, $s$
and the other problem parameters.\vspace*{-2pt}
%t2
\begin{theos}
\label{ThmNecessary}
Consider problem sequences indexed by $(n,
p,B^*,\Sigma)$ that satisfy assumptions \textup{(A1)--(A2)}, and with
minimum value $b^*_{\min}$ such that ${b^*_{\min}}^2=\Omega(\frac{\log p}{n})$, and suppose that we solve the multivariate group
Lasso with any positive regularization sequence $\{\lambda_n\}$.
Then there exist
$\nu> 0$ and
universal constants $c_i > 0$ such that if the sample
size is lower bounded as
\begin{eqnarray}
\label{EqnLowerBound}
\theta_{\ell_1/\ell_2}(n, p; B^*)
= \frac{n}{2\psi(B^*) \log(p-s)}\nonumber
 < (1-\nu)\frac{\rho_{\ell}(\Sigma_{S^c S^c | S})}{(2-\mutinc)^2},%\nonumber
\end{eqnarray}
then with probability greater than $1- c_0\exp\{-c_1 \min
(\frac{Kn}{s}, \frac{\theta_\ell}{2}
\log(p-s) ) \}$, there is no solution $\widehat{B}$
of the multivariate group Lasso that has the correct row support and
satisfies the bound $\| \widehat{B} - B^* \|_{\ell_{\infty} /\ell_{2}} =
o ( b^*_{\min})$.
\end{theos}

The proof of this claim is provided in
Section~\ref{SecProofThmNecessary}. We note that
information-theoretic methods~[\citet{Wainwright09info}] imply
that no
method (including the multivariate group Lasso) can perform exact
support recovery unless $n/s\rightarrow+\infty$, so
that the probability given in Theorem~\ref{ThmNecessary} converges to
one under the given conditions. Note that Theorems~\ref{ThmMain} and~\ref{ThmNecessary} in conjunction imply that the rescaled
sample size $\theta_{\ell_1/\ell_2}(n, p; B^*) =
\frac{n}{2 \psi(B^*) \log(p-s)}$ captures
the behavior of the multivariate group Lasso for support recovery and
estimation in block $\ell_\infty/\ell_2$ norm. For the special case
of random design matrices drawn from the standard Gaussian ensemble
(i.e., $\Sigma= I_{p\times p}$), the given scalings are
sharp:
%cor1
\begin{cors}
For the standard Gaussian ensemble, the multivariate group Lasso
undergoes a sharp
threshold at the level $\theta_{\ell_1/\ell_2}(n, p, B^*) = 1$. More
specifically, for any $\delta> 0$:
\begin{enumerate}[(a)]
\item[(a)] For problem sequences $(n, p, B^*)$ such that
$\theta_{\ell_1/\ell_2}(n, p, B^*) > 1 + \delta$, the multivariate
group Lasso succeeds
with high probability.
\item[(b)] Conversely, for sequences such that $\theta_{\ell_1/\ell_2}(n,
p, B^*) < 1 - \delta$, the multivariate group Lasso fails
with high
probability.
\end{enumerate}
\end{cors}

\begin{pf}
In the special case $\Sigma= I_{p\times p}$, it is
straightforward to verify that all the assumptions are satisfied: in
particular, we have $C_{\min}= C_{\max}= 1$, $D_{\max}= 1$ and
$\mutinc=
1$. Moreover, a short calculation shows that $\rho_{u }(I) =
\rho_{\ell}(I) = 1$. Consequently, the thresholds given in the
sufficient condition~(\ref{EqnUpperBound}) and the necessary
condition~(\ref{EqnLowerBound}) are both equal to one.
\end{pf}

%s2.2 ###
\subsection{Efficient estimation of individual supports}
\label{sec:individual-supports}

The preceding results address exact recovery of the \textit{support
union} of the regression matrix $B^*$. As demonstrated by the\vadjust{\goodbreak}
following procedure and the associated corollary of
Theorem~\ref{ThmMain}, once the row support has been recovered, it is
straightforward to recover the individual supports of each column of
the regression matrix via the additional steps of performing ordinary
least squares and thresholding.

\textit{Efficient multi-stage estimation of individual supports}:
\begin{enumerate}[(1)]
\item[(1)] estimate the \textit{support union} with $\hat{S}$, the
support union of the solution $\widehat{B}$ of the multivariate group Lasso;
\item[(2)] compute the restricted ordinary least squares (ROLS) estimate,
%
%
%e20 ###
\begin{eqnarray}
\label{EqnDefnROLS}
\widetilde{B}_{\hat{S}} & :=& \arg\min_{B_{\hat{S}}}
 | \! | \! | Y-X_{\hat{S}}B_{\hat{S}}  | \!  | \!
 |_{{F}}
\end{eqnarray}
for the restricted multivariate problem;
\item[(3)] compute the matrix $T(\widetilde{B}_{\hat{S}})$ obtained by
thresholding $\widetilde{B}_{\hat{S}}$ at the level\break $2\sqrt{\frac{2 \log(K|\hat{S}|)}{C_{\min}n}}$, and
estimate the individual supports by the nonzero entries of
$T(\widetilde{B}_{\hat{S}})$.
\end{enumerate}

The following result, which is proved in Appendix~\ref{AppIndividual},
shows that under the assumptions of Theorem~\ref{ThmMain},
the additional post-processing applied to the support union
estimate will recover the individual supports with high probability:
%cor2
\begin{cors}
\label{CorIndividual}
Under assumptions \textup{(A1)--(A3)} and the additional assumptions of
Theorem~\ref{ThmMain},
if for all individual nonzero coefficients $\beta^*_{ik}, i \in
S, 1 \leq k \leq K,$ we have $|\beta^*_{ik}|
\geq
2 \sqrt{\frac{4 \log(Ks)} {C_{\min}n}}$, then with
probability greater than $1 - \Theta(\exp(-c_0 K\log s
))$ the above
two-step estimation procedure recovers the individual supports of
$B^*$.
\end{cors}

%s2.3 ###
\subsection{\texorpdfstring{Some consequences of Theorems~\protect\ref{ThmMain}
and~\protect\ref{ThmNecessary}}{Some consequences of Theorems~1 and~2}}
\label{consequences}

We begin by making some simple observations about the sparsity-overlap
function.
%lem1
\begin{lems}\label{LemSparseOverlap}
\textup{(a)} For any design satisfying assumption \textup{(A1)}, the
sparsity-overlap $\psi(B^*)$ obeys the bounds
%
%
%e21 ###
\begin{equation}
\frac{s}{C_{\max}K} \leq \psi(B^*)
\leq
\frac{s}{C_{\min}};
\end{equation}
\smallskipamount=0pt
\begin{enumerate}
\item[(b)] if $\Sigma_{SS}=I_{s\times
s}$, and if the columns $(Z^{(k)*})$ of the matrix
$\bZ^*=\zeta(B^*)$ are orthogonal, then the sparsity-overlap
function is $\psi(B^*)=\max_{k=1, \ldots, K}
\|Z^{(k)*}\|_2^2$.
\end{enumerate}
\end{lems}

The proof of this claim is provided in
Appendix~\ref{AppLemSparseOverlap}. Based on this lemma, we now study
some special cases of Theorems~\ref{ThmMain} and~\ref{ThmNecessary}.
The simplest special case is the univariate regression problem
($K= 1$), in which case the quantity $\zeta(\beta^*)$ [as
defined in equation~(\ref{EqnDefnZ})] simply yields an
$s$-dimensional sign vector with elements
$z^*_i=\operatorname{sign}(\beta^*_i)$. [Recall that the $\operatorname
{sign}$ function is
defined as $\operatorname{sign}(0) = 0$, $\operatorname{sign}(x) = 1$
if $x>0$ and
$\operatorname{sign}(x) = -1$ if $x<0$.] In this case, the sparsity-overlap
function is given by $\psi(\beta^*) = {z^*}^T
(\Sigma_{SS})^{-1} z^*$, and as a consequence of
Lemma~\ref{LemSparseOverlap}(a), we have $\psi(\beta^*) =
\Theta(s)$. Consequently, a simple corollary of
Theorems~\ref{ThmMain} and~\ref{ThmNecessary} is that the Lasso
succeeds once the ratio $n/(2 s\log(p- s))$
exceeds a certain critical threshold, determined by the eigenspectrum
and incoherence properties of $\Sigma$, and it fails below a certain
threshold. This result matches the necessary and sufficient
conditions established in previous work on the
Lasso~[\citet{Wainwright09}].

We can also use Lemma~\ref{LemSparseOverlap} and
Theorems~\ref{ThmMain} and~\ref{ThmNecessary} to compare the
performance of the multivariate group Lasso to the following (arguably
naive) strategy for row selection using the ordinary Lasso.\vspace*{10pt}

\noindent\textit{Row selection using ordinary Lasso}:
\begin{enumerate}[(1)]
\item[(1)] Apply the ordinary Lasso separately to each of the $K$
univariate regression problems specified by the columns of $B^*$,
thereby obtaining estimates $\widehat{\beta}^{(k)}$ for $k=1, \ldots,
K$.
\item[(2)] For $k=1, \ldots, K$, estimate the support of individual
columns via
$\widehat{S}_k :=\{i \mid \widehat{\beta}^{(k)}_i \neq0 \}$.
\item[(3)] Estimate the row support by taking the union: $\widehat{S} =
\bigcup_{k=1}^K\widehat{S}_k$.
\end{enumerate}
To understand the conditions governing the success/failure of this
procedure, note that it succeeds if and only if for each nonzero row
$i \in S= \bigcup_{k=1}^KS_k$, the variable
$\widehat{\beta}^{(k)}_i$ is nonzero for at least one $k$, and for all
$j \in{S^c}= \{1, \ldots, p\} \backslash S$, the variable
$\widehat{\beta}^{(k)}_j = 0$ for all $k = 1, \ldots, K$. From
our understanding of the univariate case, we know that for
% MJ \contcrit hasn't been defined. And shouldn't this be \theta(n, p,
%B^*)
% in any case?
$\theta_u=\frac{\rho_{u }(\Sigma_{S^c S^c | S
})}{\mutinc^2}$, the condition
%
%
%e22 ###
\begin{equation}
\hspace*{30pt}n \geq 2 \theta_u \max_{k = 1, \ldots,
K}
\psi\bigl(\beta^{*(k)}_S\bigr) \log(p- s_k) \geq 2 \theta_u
\max
_{k = 1, \ldots, K} \psi\bigl(\beta^{*(k)}_S\bigr)
\log(p- s)
\end{equation}
is sufficient to ensure that the ordinary Lasso succeeds in row
selection. Conversely, for $\theta_\ell=\frac{\rho_{\ell}(\Sigma
_{S^c S^c | S})}{(2-\mutinc)^2}$, if the sample size is
upper bounded as
\[
n< 2 \theta_{\ell} \max_{k = 1, \ldots, K} \psi
\bigl(\beta^{*(k)}_S\bigr)
\log(p- s),
\]
then there will exist some $j \in{S^c}$ such for at least one $k \in
\{1, \ldots, K\}$, there holds $\widehat{\beta}^{(k)}_j \neq0$
with high probability, implying failure of the ordinary Lasso.

A natural question is whether the multivariate group Lasso, by taking
into account
the couplings across columns, always outperforms (or at least matches)
the naive strategy. The following result, proven in
Appendix~\ref{AppCorGroupOrd}, shows that if the design is
uncorrelated on the support, then indeed this is the case.
%cor3
\begin{cors}[(Multivariate group Lasso versus ordinary Lasso)]\label{CorGroupOrd}
Assume $\Sigma_{S
S} = I_{s\times s}$. Then for any multivariate
regression problem, row selection using the ordinary Lasso strategy
requires, with high probability, at least as many samples as the
$\ell_1/\ell_2$ multivariate group Lasso. In particular, the relative
efficiency of
multivariate group Lasso versus ordinary Lasso is given by the ratio
%
%
%e23 ###
\begin{eqnarray}
\frac{ \max_{k=1, \ldots, K} \psi(\beta_S^{*(k)}) \log
(p- s_k)}{\psi(B^*_S) \log(p- s)} &
\geq
& 1.
\end{eqnarray}
\end{cors}

We consider the special case of identical regressions, for
which a result can be stated for any covariance design
and the case of ``orthonormal'' regressions which illustrates
Corollary~\ref{CorGroupOrd}.
%
%ex1
\begin{exas}[(Identical regressions)]
Suppose that $B^*:=\beta^*\vec{1}_K^{ T}$---that is,
$B^*$ consists of $K$ copies of the same coefficient vector
$\beta^*\in\mathbb{R}^p$, with support of cardinality $|S|
= s$. We then have $[\zeta(B^*)]_{ij} =
\operatorname{sign}(\beta^*_{i})/\sqrt{K}$, from which we see that
$\psi(B^*) = {z^*}^T (\Sigma_{SS})^{-1} z^*$, with
$z^*$ being an $s$-dimensional sign vector with elements
$z^*_i=\operatorname{sign}(\beta^*_i)$. Consequently, we have the equality
$\psi(B^*) =\psi(\beta^{*})$, so that under our analysis
there is no benefit in using the multivariate group Lasso relative to
the strategy of solving separate Lasso problems and constructing the
union of individually estimated supports. This behavior may seem
counter-intuitive, since under the model~(\ref{EqnGenObsModel}) we
essentially have $Kn$ observations of the coefficient
vector $\beta^*$ with the same design matrix but $K$ independent
noise realizations, which could help to reduce the effective noise
variance from $\sigma^2$ to $\sigma^2/K$ if the fact that the
regressions are identical is known. It must be borne in mind,
however, that in our high-dimensional analysis the noise variance does
not grow as the dimensionality grows, and thus asymptotically the
noise is dominated by the interference between the covariates, which
grows as $(p- s)$. It is thus this high-dimensional
interference that dominates the rates given in Theorems~\ref{ThmMain}
and~\ref{ThmNecessary}.
\end{exas}

In contrast to this pessimistic example, we now turn to the
most optimistic extreme:
%ex2
\begin{exas}[(``Orthonormal'' regressions)]
\label{ExaOrtho}
Suppose that $\Sigma_{SS} = I_{s\times s}$
and (for $s> K$) suppose that $B^*$ is constructed
such that the columns of the $s\times K$ matrix
$\zeta(B^*)$ are orthogonal and with equal norm (which implies
their norm equals $\sqrt{\frac{s}{K}}$). Under these
conditions, we
claim that the sample complexity of multivariate group Lasso is smaller
than that of
the ordinary Lasso by a factor of $1/K$. Indeed, using
Lemma~\ref{LemSparseOverlap}(b), we observe that
\[
K\psi(B^*)= K\big\|Z^{(1)*}\big\|^2 =\sum_{k=1}^K
\big\|Z^{(k)*}\big\|^2=\operatorname{tr}({\bZ^*}^T\bZ^*)=\operatorname{tr}
(\bZ^*{\bZ^*}^T )=s,
\]
because $\bZ^*{\bZ^*}^T \in\mathbb{R}^{s \times s}$ is the Gram
matrix of
$s$ unit vectors in $\mathbb{R}^{k}$ and its diagonal elements are therefore
all equal to $1$. Consequently, the multivariate group Lasso recovers
the row
support with high probability for sequences such that
\begin{eqnarray*}
\frac{n}{2 \sklfrac{s}{K} \log(p-
s)}
& > & 1,
\end{eqnarray*}
which allows for sample sizes $K$ times smaller than the ordinary
Lasso approach.
\end{exas}

Corollary~\ref{CorGroupOrd} and the subsequent examples address
the case of uncorrelated design ($\Sigma_{SS} = I_{s
\times s}$) on the row support $S$, for which the
multivariate group Lasso is never worse than the ordinary Lasso in
performing row
selection. The following example shows that if the supports are
disjoint, the ordinary Lasso has the same sample complexity as the
multivariate group Lasso for uncorrelated design $\Sigma_{SS
} =
I_{s\times s}$, but can be better than the multivariate
group Lasso
for designs $\Sigma_{SS}$ with suitable correlations:
%cor4
\begin{cors}[(Disjoint supports)]
\label{CorDisSupport}
Suppose that the support sets $S_k$ of individual regression
problems are disjoint. Then for any design covariance
$\Sigma_{SS}$, we have
%
%
%e24 ###
\begin{eqnarray}
\label{EqnDisSupport}
\max_{1 \leq k \leq K} \psi\bigl(\beta_S^{(k)*}\bigr)
\stackrel{\mathrm{(a)}}{\leq}  \psi(B^*_S)  \stackrel{\mathrm{(b)}}{\leq}
\sum_{k=1}^K\psi\bigl(\beta_S^{(k)*}\bigr).
\end{eqnarray}
\end{cors}

\begin{pf}
First note that, since all supports are disjoint,
$Z^{(k)*}_{i}=\operatorname{sign}
(\beta^*_{ik})$,
so that $Z^{(k)*}_S=\zeta(\beta^{(k)*}_S)$. Inequality (b)
is then immediate because we have $ | \!  | \! | {\bZ
^*_S}^T\Sigma_{S S}^{-1}{\bZ^*_S}  | \!  | \!  |_{{2}}{}
\leq\operatorname{tr}({\bZ^*_S}^T\Sigma_{S
S}^{-1}{\bZ^*_S})$. To establish inequality (a), we note that
\begin{eqnarray*}
\psi(B^*) &=& \max_{x \in\mathbb{R}^{K} : \|x\| \leq1}
x^T{\bZ^*_S}^T\Sigma_{SS}^{-1}{\bZ^*_S}x \geq \max_{1
\leq k
\leq K} e_k^T{\bZ^*_S}^T\Sigma_{SS}^{-1}{\bZ^*_S}
e_k\\ & \geq& \max_{1 \leq k \leq K} {Z^{(k)*}_S
}^T\Sigma_{S
S}^{-1}{Z^{(k)*}_S}.
\end{eqnarray*}
\upqed
\end{pf}

A caveat in interpreting Corollary~\ref{CorDisSupport}, and more
generally in comparing the performance of the ordinary Lasso and the
multivariate group Lasso, is that for a general covariance matrix
$\Sigma_{SS}$, assumptions (A2) and (A3) required
by Theorem~\ref{ThmMain} do not induce the same constraints on the
covariance matrix $\Sigma$ when applied to the multivariate problem as
when applied to the individual regressions. Indeed,\vspace*{-2pt} in the latter
case, (A2) would require $\max_k  | \! | \! | \Sigma
_{S_k^c S_k} \Sigma_{S_kS_k}^{-1}  | \!  | \!  |_{{\infty
}}\leq1-\mutinc$ and (A3) would require
$\max_k  | \! | \! | \Sigma_{S_kS_k}^{-1}  | \!
| \!  |_{{\infty}} \leq D_{\max}$. Thus
(A3) is a stronger assumption in the multivariate case but (A2) is
not.

We illustrate Corollary~\ref{CorDisSupport} with an example.
%ex3
\begin{exas}[(Disjoint support with uncorrelated design)]
Suppose that\break\mbox{$\Sigma_{SS}=I_{s\times s}$}, and the supports are disjoint. In this case, we claim
that the sample complexity of the $\ell_1/\ell_2$ multivariate group
Lasso is the
same as the ordinary Lasso. If the individual regressions have
disjoint support, then ${\bZ^*_S}=\zeta(B_S^*)$ has only a single
nonzero entry per row and therefore the columns of $Z^*$ are
orthogonal. Moreover, $Z^*_{ik}=\operatorname{sign}(\beta
^{(k)*}_{i})$. By
Lemma~\ref{LemSparseOverlap}(b), the sparsity-overlap function
$\psi(B^*)$ is equal to the largest squared column norm. But
$\|Z^{(k)*}\|^2=\sum_{i=1}^s
\operatorname{sign}(\beta^{(k)*}_{i})^2=s_k$. Thus, the sample
complexity of
the multivariate group Lasso is the same as the ordinary Lasso in this
case.\footnote{In making this assertion, we are ignoring any
difference between $\log(p- s_k)$ and $\log(p- s)$,
which is valid, for instance, in the regime of sublinear sparsity,
when $s_k/p\rightarrow0$ and $s/p\rightarrow0$.}
\end{exas}

 Finally, we consider an example that illustrates the effect
of correlated designs:

%ex4
\begin{exas}[(Effects of correlated designs)]
\label{twobytwo}
To illustrate the behavior of the sparsity-overlap function in the
presence of correlations in the design, we consider the simple case
of two regressions with support of size 2. For parameters
$\vartheta_1$ and $\vartheta_2 \in[0, \pi]$ and $\mu\in(-1,
+1)$, consider regression matrices $B^*$ such that
$B^*=\zeta(B^*_S)$ and
%
%
%e25 ###
\begin{equation}
\zeta(B^*_S)=
\left[
\matrix{
\cos(\vartheta_1) & \sin(\vartheta_1) \vspace{2pt}\cr
\cos(\vartheta_2) & \sin(\vartheta_2)
}
\right]
\quad\mbox{and}\quad \Sigma_{SS}^{-1}=
\left[
\matrix{
1 & \mu\vspace{2pt}\cr \mu& 1
}
\right]
.
\end{equation}
Setting $M^*={\zeta(B^*_S)}^T \Sigma_{S
S}^{-1}\zeta(B^*_S)$, a simple calculation shows that
\[
\operatorname{tr}(M^*) = 2 \bigl(1+\mu\cos(\vartheta_1 - \vartheta_2)\bigr)
\quad\mbox{and}\quad \det(M^*) =
(1-\mu^2)\sin(\vartheta_1 - \vartheta_2)^2,
\]
so that the eigenvalues of $M^*$ are
\[
\mu^+ = (1+\mu) \bigl(1+\cos(\vartheta_1 - \vartheta_2)\bigr)
\quad\mbox{and}\quad \mu^- = (1-\mu)
\bigl(1-\cos(\vartheta_1 - \vartheta_2)\bigr),
\]
and $\psi(B^*)=\max(\mu^+,\mu^-)$. On the other hand, with
\begin{eqnarray*}
\tilde{z}_1=\zeta\bigl(\beta^{(1)*}\bigr)=
\pmatrix{\operatorname{sign}(\cos(\vartheta_1)) \vspace{2pt}\cr
\operatorname{sign}(\cos(\vartheta_2))
}
\quad\mbox{and}\quad
\tilde{z}_2 = \zeta\bigl( \beta^{(2)*}\bigr) =
\pmatrix{\operatorname{sign}(\sin(\vartheta_1)) \vspace{2pt}\cr
\operatorname{sign}(\sin(\vartheta_2))
}
\end{eqnarray*}
we have
\begin{eqnarray*}
\begin{aligned}
\psi\bigl(\beta^{(1)*}\bigr) =& \tilde{z}_1^T \Sigma_{SS}^{-1}
\tilde{z}_1 = \mathbf{1}_{\{\cos(\vartheta_1) \neq0\}}+\mathbf
{1}_{\{\cos(\vartheta_2) \neq0\}}+2 \mu\operatorname{sign}(\cos
(\vartheta_1)\cos(\vartheta_2)),\\
\psi\bigl(\beta^{(2)*}\bigr) =& \tilde{z}_2^T \Sigma_{SS}^{-1}
\tilde{z}_2 = \mathbf{1}_{\{\sin(\vartheta_1) \neq0\}}+\mathbf
{1}_{\{\sin(\vartheta_2) \neq0\}}+2 \mu\operatorname{sign}(\sin
(\vartheta_1)\sin(\vartheta_2)).
\end{aligned}
\end{eqnarray*}

%f1 ###
\begin{figure}

\includegraphics{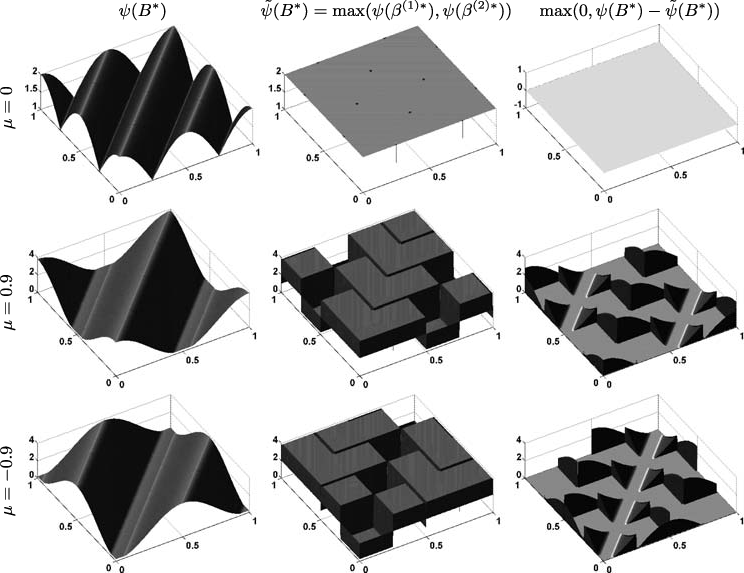}

\caption{Comparison of sparsity-overlap functions
for $\ell_1/\ell_2$ and the Lasso. For the pair $\frac{1}{2\pi
}(\vartheta_1,\vartheta_2)$, we represent in each row of plots,
corresponding respectively to $\mu=0$ (top), $0.9$ (middle) and $-0.9$
(bottom), from left to right, the quantities: $\psi(B^*)$ (left),
$\max(\psi(\beta^{(1)*}),\psi(\beta^{(2)*}))$ (center) and $\max
(0,\psi(B^*)-\max(\psi(\beta^{(1)*}),\psi (\beta^{(2)*})))$
(right). The latter indicates when the inequality $\psi(B^*) \leq\max
(\psi(\beta^{(1)*}),\psi(\beta^{(2)*}))$ does not hold and by how
much it is violated.}\label{FigSurface}
\end{figure}

Figure~\ref{FigSurface} provides a graphical comparison of these
sample complexity functions. The function
$\tilde{\psi}(B^*)=\max(\psi(\beta^{(1)*}),\psi
(\beta^{(2)*}))$
is discontinuous on $\mathcal{S}=\frac{\pi}{2}\mathbb{Z}\times
\mathbb{R}
\cup
\mathbb{R}\times \frac{\pi}{2}\mathbb{Z}$, and, as a consequence,
so is its
difference with $\psi(B^*)$. Note that, for fixed
$\vartheta_1$ or fixed $\vartheta_2$, some of these discontinuities
are \textit{removable discontinuities} of the induced function on the
other variable, and these discontinuities therefore create needles,
slits or flaps in the graph of the function $\tilde{\psi}$. Denote
by $\mathcal{R}^+$ and $\mathcal{R}^-$ the sets
\begin{eqnarray*}
\mathcal{R}^+=\{ (\vartheta_1,\vartheta_2) |
\min[\cos(\vartheta_1)\cos(\vartheta_2),
\sin(\vartheta_1)\sin(\vartheta_2)]>0 \},
\\
\mathcal{R}^- = \{ (\vartheta_1,\vartheta_2) |
\max[\cos(\vartheta_1)\cos(\vartheta_2)
,\sin(\vartheta_1)\sin(\vartheta_2)]<0 \}
\end{eqnarray*}
on which $\tilde{\psi}(B^*)$ reaches its minimum value
when $\mu\geq0.5$ and $\mu\leq0.5$, respectively (see
middle and bottom center plots in Figure~\ref{FigShiftsToeplitz}). %\ref{surfs}).
For $\mu=0$, the top center graph illustrates that
$\tilde{\psi}(B^*)$ is equal to $2$ except for the cases of
matrices $B^*_S$ with disjoint support, corresponding to the
discrete set $\mathcal{D}=\{(k\frac{\pi}{2},(k\pm1)\frac{\pi
}{2}), k
\in\mathbb{Z}\}$ for which it equals $1$. The top rightmost graph
illustrates that, as shown in Corollary~\ref{CorGroupOrd}, the
inequality always holds for an uncorrelated design. For $\mu> 0$,
the inequality $\psi(B^*) \leq
\max(\psi(\beta^{(1)*}),\psi(\beta^{(2)*}))$ is violated only
on a subset of $\mathcal{S} \cup\mathcal{R}^-$; and for $\mu< 0$,
the inequality is symmetrically violated on a subset of $\mathcal{S}
\cup\mathcal{R}^+$ (see Figure~\ref{FigShiftsToeplitz}). %\figref{surfs}).
\end{exas}
%

%s2.4 ###
\subsection{Illustrative simulations}

In this section, we present the results of simulations that
illustrate the sharpness of Theorems~\ref{ThmMain}
and~\ref{ThmNecessary}, and furthermore demonstrate how quickly the
predicted behavior is observed as elements of the triple $(n$,
$p$, $s)$ grow in different regimes. We explore the case
of two regressions (i.e., $K= 2$) which share an identical
support set $S$ with cardinality $|S| = s$ in
Section~\ref{phas_trans} and consider a slightly more general case in
Section~\ref{emp_thres}.

%s2.4.1 ###
\subsubsection{Threshold effect in the standard Gaussian case}
\label{phas_trans}

%f2 ###
\begin{figure*}

\includegraphics{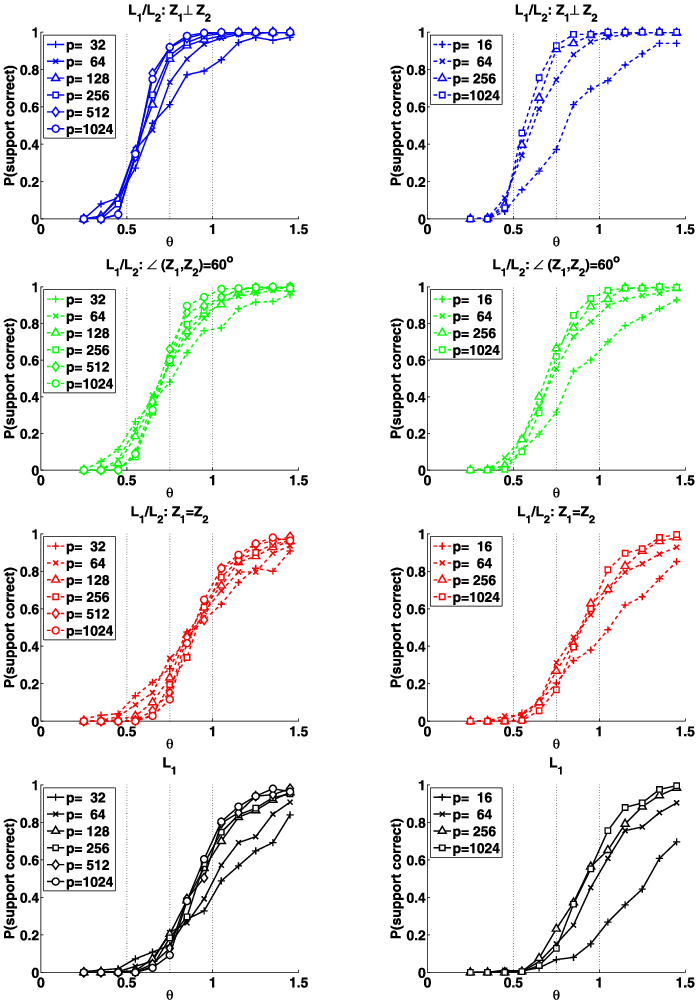}

\caption{Plots of support union recovery probability, $\mathbb
{P}[\widehat{S} = S]$, versus the control parameter $\theta= n/[2
s\log(p- s)]$ for two different types of sparsity: linear sparsity in
the left column ($s = p/8$) and logarithmic sparsity in the right
column ($s = 2\log_2(p))$. The first three rows are based on using
the multivariate group Lasso to estimate the support for the three
cases of identical regression, intermediate angles and orthonormal
regressions. The fourth row presents results for the Lasso in the case
of identical regressions.} \label{FigDiffP}
\end{figure*}
The first set of experiments was designed to reveal the threshold
effect predicted by Theorems~\ref{ThmMain} and~\ref{ThmNecessary}.
The design matrix $\bX $ is sampled from the standard Gaussian
ensemble, with i.i.d.\ entries $X_{ij} \sim N(0,1)$. We consider two
types of sparsity:
\begin{itemize}
\item logarithmic sparsity, where $s = \alpha\log(p)$, for $\alpha =
2/\log(2)$, and
\item linear sparsity, where $s = \alpha p$, for $\alpha= 1/8$
\end{itemize}
for various ambient model dimensions $p\in\{16, 32, 64, 256,512,1024\}
$. For a given triplet $(n, p, s)$, we solve the block-regularized
problem~(\ref{EqnBlockRegProb}) with the regularization parameter
$\lambda_n= \sqrt{{ \log (p-s) (\log s)}/{n}}$. For each fixed $(p,
s)$ pair, we measure the sample complexity in terms of a parameter
$\theta$, in particular letting $n= \theta s\log(p- s)$ for
$\theta\in[0.25, 1.5]$. We let the matrix $B^*\in\mathbb
{R}^{p\times2}$ of \mbox{regression} coefficients have entries $\beta
^*_{ij}$ in $\{-1/\sqrt{2}, 1/\sqrt{2}\}$, choosing the parameters to
vary the angle between the two columns, thereby obtaining various
desired values of $\psi(B^*)$. Since $\Sigma=I_{p\times p}$ for the
standard Gaussian ensemble, the sparsity-overlap function $\psi(B^*)$
is simply the maximal eigenvalue of the Gram matrix $\zeta(B^*_S)^T
\zeta(B^*_S)$. Since $|\beta^*_{ij}|= 1/\sqrt{2}$ by construction,
we are guaranteed that $B^*_S = \zeta(B^*_S)$, that the minimum value
$b^*_{\min}= 1$, and, moreover, that the columns of $\zeta(B^*_S)$ have
the same Euclidean norm.

To construct parameter matrices $B^*$ that
satisfy $|\beta^*_{ij}|={1}/{\sqrt{2}}$, we choose both $p$ and the
sparsity scalings so that the obtained values for $s$ are multiples of
four. We then construct the columns $Z^{(1)*}$ and $Z^{(2)*}$ of the
matrix $B^*=\zeta(B^*)$ from copies of vectors of length four.
Denoting by $\otimes$ the usual matrix tensor product, we consider the
following $4$-vectors:
\begin{description}
\item[Identical regressions:] We set $Z^{(1)*} = Z^{(2)*} = \frac
{1}{\sqrt{2}} \vec{1}_s$, so that the sparsity-overlap function is
$\psi(B^*) = s$.
\item[Orthonormal regressions:] Here $B^*$ is constructed with
$Z^{(1)*} \perp Z^{(2)*}$, so that $\psi(B^*) = \frac{s}{2}$, the
most favorable situation. In order to achieve this orthonormality, we
set $Z^{(1)*} =\frac{1}{\sqrt{2}} \vec{1}_s$ and $Z^{(2)*} =\frac
{1}{\sqrt{2}} \vec{1}_{s/2} \otimes(1,-1)^T$.
\item[Intermediate angles:] In this intermediate case, the columns
$Z^{(1)*}$ and $Z^{(2)*}$ are at a $60^\circ$ angle, which leads to
$\psi(B^*) = \frac{3}{4} s$. Specifically, we set $Z^{(1)*} =\frac
{1}{\sqrt{2}} \vec{1}_s$ and $Z^{(2)*} =\frac{1}{\sqrt{2}}
\vec{1}_{s/4} \otimes (1,1,1,-1)^T$.
\end{description}
%
%f3 ###
\begin{figure*}

\includegraphics{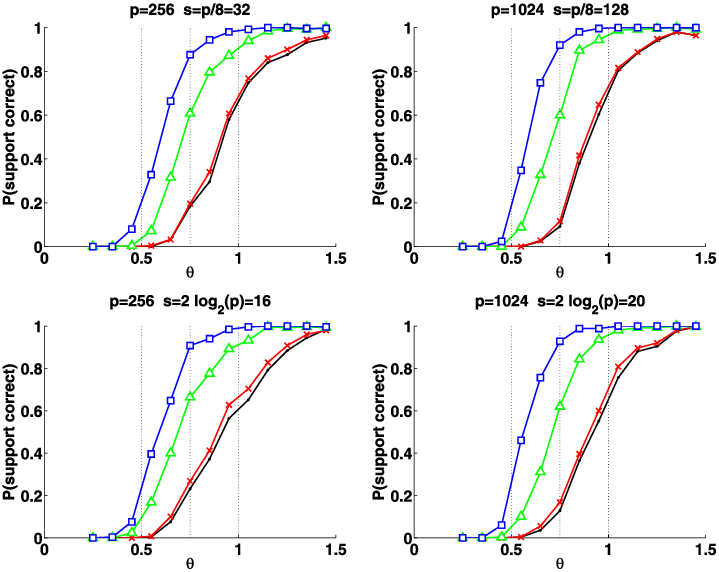}

\caption{Plots of support recovery probability, ${\mathbb{P}[\widehat
{S} = S]}$, versus the control parameter $\theta= n/[2 s\log(p- s)]$
for two different type of sparsity: logarithmic sparsity on top ($s =
\mathcal{O}(\log(p))$) and linear sparsity on bottom ($s = \alpha
p$), and for increasing values of $p$ from left to right. The noise
level is set at $\sigma=0.1$. Each graph shows four curves (black,
red, green, blue) corresponding to the case of independent $\ell_1$
regularization, and, for $\ell_1/\ell_2$ regularization, the cases of
identical regression, intermediate angles and orthonormal regressions.
Note how curves corresponding to the same case across different problem
sizes $p$ all coincide, as predicted by Theorems~\protect\ref{ThmMain}
and~\protect\ref{ThmNecessary}. Moreover, consistent with the theory, the
curves for the identical regression group reach $\mathbb{P}[\widehat
{S} = S] \approx0.50$ at $\theta\approx 1$, whereas the orthonormal
regression group reaches $50\%$ success substantially earlier.} \label{FigShifts}
\end{figure*}
%
%% END NONFIG

Figure~\ref{FigDiffP} shows plots for linear sparsity (left column)
and logarithmic sparsity (right column) in all three cases and where
the multivariate group Lasso was used (top three rows), as well as
the reference Lasso case for the case of identical regressions (bottom
row). Each panel plots the success probability, $\mathbb{P}[\widehat{S}
= S]$, versus the rescaled sample size $\theta= n/[2
s\log(p- s)]$. Under this rescaling,
Theorems~\ref{ThmMain} and~\ref{ThmNecessary} predict that the curves
should align, and that the success probability should transition to
$1$ once $\theta$ exceeds a critical threshold (dependent on the type
of ensemble). Note that for suitably large problem sizes ($p\geq
128$), the curves do align in the predicted way, showing step-function
behavior. Figure~\ref{FigShifts} plots data from the same simulations
in a different format. Here the top row corresponds to logarithmic
sparsity, and the bottom row to linear sparsity; each panel shows the
four different choices for $B^*$, with the problem size $p$
increasing from left to right. Note how in each panel the location of
the transition of $\mathbb{P}[\widehat{S} = S]$ to one shifts from
right to left, as we move from the case of identical regressions to
intermediate angles to orthogonal regressions.

%s2.4.2 ###
\subsubsection{Threshold effect with Toeplitz covariance matrices}

%%%%%%%%%%%%%%%%

The simulations in the previous section involved the standard Gaussian
design matrix; in this section, we explore the behavior for design
matrices with some dependence structure. In particular, we report
results for random designs with rows drawn from a Gaussian with
Toeplitz covariance matrix of the form $\Sigma= (\rho^{|i-j|}
)_{1 \leq i,j \leq p}$, for some parameter $\rho\in[0,1)$.
\citet{Zhao06} have shown that such Toeplitz matrices satisfy the
%mutual incoherence
irrepresentable
conditions required for support consistency. As
with our experiments in the standard Gaussian case, we consider the
same two regimes (linear and logarithmic), using the same families of
regression matrices $B^*$ and the same noise level. We select
the support of the regression matrices as a random subset of the
$p$ covariates of size $s$, and draw the design matrices
from the Toeplitz ensemble $\rho=0.5$. For each pair
$(s,p)$, we consider a number of observations of the form
$n=2 \theta s\log(p)$ for $\theta\in
[0.25,4]$.
%Note that for the matrices considered, the mutual
%incoherence condition is satisfied for a parameter of mutual
%incoherence $\mutinc$ which varies with the dimension $\pdim$;
%similarly the smallest eigenvalue of $\XX$ is also changing with
%$\spindex$.
%However, for the reasons outlined
%above, the transitions are not perfectly aligned; moreover, unlike the
%standard Gaussian Case, there is a gap between the thresholds
%corresponding to necessary and sufficient conditions and as a result
%the phase transition should be expected to be less sharp.
%% BEGIN COMMENT
%%
%%\begin{figure*}
%{P}[\widehat{S} = S]$ versus the control parameter $\theta= n/[2
%s\log(p- s)]$ when the covariance matrix is a Toeplitz matrix with
%parameter $\rho=0.5$, for the same protocol as in Figure~\ref
%{FigDiffP}} \label{FigDiffPToeplitz}
%%
%}
%%%% END COMMENT

\begin{figure*}

\includegraphics{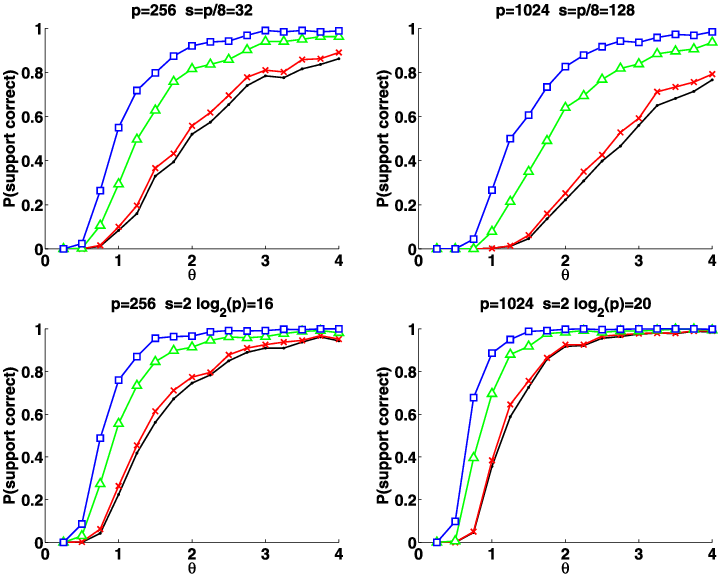}

\caption{Plots of support recovery probability, ${\mathbb{P}[\widehat
{S} = S]}$, versus the control parameter $\theta= n/[2 s\log(p- s)]$
when the covariance matrix is a Toeplitz matrix with parameter $\rho
=0.5$, for the same protocol as described in Figure~\protect\ref{FigShifts}.}
\label{FigShiftsToeplitz}
\end{figure*}

Figure~\ref{FigShiftsToeplitz} is the analog of the previously shown
Figure~\ref{FigShifts}: for problems with random designs from the
Toeplitz ensemble, it plots the support recovery probability
${\mathbb{P}[\widehat{S} = S]}$ versus the control parameter
$\theta= n/[2 s\log(p- s)]$ for two
different types of sparsity---logarithmic sparsity on top ($s =
\mathcal{O}(\log(p))$) and linear sparsity on bottom ($s = \alpha
p$). The four curves (black, red, green, blue) corresponding to the
case of independent $\ell_1$ regularization, and, for $\ell_1/\ell_2$
regularization, the cases of identical regression, intermediate
angles and orthonormal regressions. Qualitatively, note that we
observe the same type of transitions as in the standard
Gaussian case; moreover, the curves shift from right to left as the
angles between the regression columns vary from orthogonal to
identical.

%f4 ###

%s2.4.3 ###
\subsubsection{Empirical threshold values}
\label{emp_thres}

In this experiment, we aim at verifying more precisely the location of
the $\ell_1/\ell_2$ threshold as the regression vectors vary
continuously from identical to orthogonal with equal length. We
consider the case of
matrices $B^*$ of size $s\times2$ for $s$ even.
In Example~\ref{twobytwo} of Section~\ref{consequences}, we
characterized the value of $\psi(B^*)$ when $B^*$ is a $2
\times2$ matrix.

In order to generate a family of regression matrices with smoothly
varying sparsity-overlap function consider the following $2 \times2$
matrix:
%
%
%e26 ###
\begin{equation}
B_1(\alpha)=
\left[
\matrix{
\displaystyle \frac{1}{\sqrt{2}} & \displaystyle \frac{1}{\sqrt{2}}\vspace{2pt}\cr
\cos\biggl(\displaystyle \frac{\pi}{4}+\alpha\biggr) & \sin\biggl(\displaystyle \frac{\pi}{4}+\alpha\biggr)
}
\right]
.
\end{equation}
Note that $\alpha$ is the angle between the two \textit{rows} of
$B_1(\alpha)$ in this setup. Note, moreover, that the columns of
$B_1(\alpha)$ have varying norm.

We use this base matrix to define the following family of
regression matrices $B^*_S \in\mathbb{R}^{s\times2}$:
%
%
%e27 ###
\begin{eqnarray}
\mathcal{B}_1 & :=& \biggl\{ B_{1s}(\alpha)=\vec{1}_{s/2}
\otimes B_1(\alpha), \alpha\in\biggl[0, \frac{\pi}{2} \biggr]
\biggr\}.
\end{eqnarray}
For a design matrix drawn from the standard Gaussian ensemble, the
analysis of Example~\ref{twobytwo} in Section~\ref{consequences}
extends naturally to show that the sparsity-overlap function is
$\psi(B_{s1}(\alpha))=\frac{s}{2}(1+|\cos(\alpha)|)$. Moreover,
as we vary $\alpha$ from $0$ to $\frac{\pi}{2}$, the two regressions
vary from identical to orthonormal and the sparsity-overlap function
decreases from $s$ to $\frac{s}{2}$.

We fix the problem size $p=2048$ and sparsity
$s=\log_2(p)=22$. For each value of $\alpha\in[0,
\frac{\pi}{2}]$, we generate a matrix from the specified family and
angle. We then solve the multivariate group Lasso optimization
problem~(\ref{EqnBlockRegProb}) with sample size $n= 2 \theta
s\log(p- s)$ for a range of values of $\theta$ in
$[0.25,1.5]$; for each value of~$\theta$, we repeat the experiment
(generating random design matrix $X$ and observation matrix $Y$ each
time) over $T = 500$ trials. Based on these trials, we then estimate
the value of $\theta_{50\%}$ for which the exact support is retrieved at
least $50\%$ of the time. Since
$\psi(B^*)=\frac{1+|\cos(\alpha)|}{2} s$, our
theory predicts that if we plot $\theta_{50\%}$ versus $|\cos(\alpha)|$,
the plot should lie on or below the straight line
$\frac{1+|\cos(\alpha)|}{2}$. We also perform the same experiments
for row selection using the ordinary Lasso and plot the resulting
estimated thresholds on the same axes.

%f5 ###
\begin{figure*}

\includegraphics{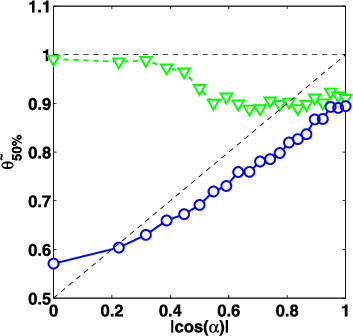}

\caption{Plots of the Lasso sample complexity $\theta= n/[2 s\log(p-
s)]$ for which the probability of support union recovery exceeds $50\%$
empirically as a function of $|\cos(\alpha)|$ for $\ell_1$-based
recovery and $\ell_1/\ell_2$-based recovery, where $\alpha$ is the
angle between ${Z^{(1)*}}$ and $Z^{(2)*}$ for the family $\mathcal
{B}_1$. We consider the two following methods for performing row
selection: Ordinary Lasso ($\ell_1$, green triangles) and multivariate
group Lasso (blue circles). } \label{Figthres}
\end{figure*}

The results are shown in Figure~\ref{Figthres}. Note first that
the curve obtained for $\widehat{S}_{\ell_1/\ell_2}$ (blue circles)
coincides roughly with the theoretical prediction,
$\frac{1+|\cos(\alpha)|}{2}$ (black dashed diagonal) as regressions vary
from orthogonal to identical. Moreover, the estimated $\theta_{50\%}$
of the ordinary Lasso remains above $0.9$ for all values of $\alpha$,
close to the theoretical value of $1$. However, the curve
obtained is not constant, but is roughly sigmoidal with a first
plateau close to $1$ for $\cos(\alpha) <0.4$ and a second plateau
close to $0.9$ for $\cos(\alpha) >0.5$. The latter coincides with
the empirical value of $\theta_{50\%}$ for the univariate Lasso for
the first column $\beta^{(1)*}$ (not shown). There are two reasons
why the value of $\theta_{50\%}$ for the ordinary Lasso does not
match the prediction of the first-order asymptotics: first, for
$\alpha=\frac{\pi}{4}$ [corresponding to $\cos(\alpha) = 0.7$],
the support of $\beta^{(2)*}$ is reduced by one half and therefore
its sample complexity is decreased in that region. Second, the
supports recovered by individual Lassos for $\beta^{(1)*}$ and
$\beta^{(2)*}$ vary from uncorrelated when $\alpha=\frac{\pi}{2}$
to identical when $\alpha=0$. It is therefore not surprising that
the sample complexity is the same as a single univariate Lasso
for $\cos(\alpha)$ large and higher for $\cos(\alpha)$ small,
where independent estimates of the support are more likely to
include, by union, spurious covariates in the row support.

%s3 ###
\section{\texorpdfstring{Proof of Theorem~\protect\ref{ThmMain}}{Proof of Theorem~1}}\label{SecProof}

In this section, we provide the proof of Theorem~\ref{ThmMain}, which
gives sufficient conditions for success of the multivariate group
Lasso. Subsequently, in Section~\ref{SecProofThmNecessary}, we
provide the proof for the necessary conditions as given in
Theorem~\ref{ThmNecessary}. For the convenience of the reader, we
begin by recapitulating the notation to be used throughout both of
these arguments:
\begin{itemize}
\item The sets $S$ and ${S^c}$ are a partition of the set of
columns of $\bX $, such that \mbox{$|S|= s$,}
$|{S^c}|= p- s$.
\item The design matrix is partitioned as $\bX = [
\bX _S \bX _{{S^c}} ]$, where $\bX _S \in\mathbb{R}^{n
\times s}$ and $\bX _{{S^c}} \in\mathbb{R}^{n\times
(p-s)}$.
\item The regression coefficient matrix is also partitioned as
$B^*=
\big[
{\fontsize{8.36}{8.36}\selectfont
\matrix{ B^*_S\cr B^*_{S^c}
}}
\big]
$, with \mbox{$B^*_S\in\mathbb{R}^{s\times K}$}
and $B^*_{{S^c}}=0 \in\mathbb{R}^{(p-s) \times K
}$. We
use $\beta^*_i$ to denote the $i$th row of $B^*$.
\item The regression model is given by $ \bY=\bX  B^*+
W$, where
the noise matrix \mbox{$W \in\mathbb{R}^{n \times K}$} has i.i.d.\ $N(0,
\sigma^2)$
entries.
\item The matrix ${\bZ^*_S}= \zeta(B^*_S) \in\mathbb{R}
^{s\times
K}$ has rows $Z^*_i=\zeta(\beta^*_i) =
\frac{\beta^*_i}{\|\beta^*_i\|_2} \in\mathbb{R}^K$.
\end{itemize}

%s3.1 ###
\subsection{High-level proof outline}

At a high level, the proof is based on the notion of a
\textit{primal--dual witness}: we construct a primal matrix $\widehat{B}$
along with a dual matrix $\widehat{\bZ}$ such that:
\begin{enumerate}[(a)]
\item[(a)] the pair $(\widehat{B}, \widehat{\bZ})$ together satisfy the
Karush--Kuhn--Tucker (KKT) conditions associated with the second-order
cone program~(\ref{EqnBlockRegProb}), and
\item[(b)] this solution certifies that the multivariate group Lasso
recovers the union of supports $S$.
\end{enumerate}
For general high-dimensional problems (with $p\gg n$), the
multivariate group Lasso of~(\ref{EqnBlockRegProb}) need not have a
unique solution; however, a consequence of our theory is that the
constructed solution $\widehat{B}$ is the unique optimal solution under
the conditions of Theorem~\ref{ThmMain}.

We begin by noting that the block-regularized
problem~(\ref{EqnBlockRegProb}) is convex, and not differentiable for
all $B$. In particular, denoting by ${\beta}_i$ the $i$th row
of $B$, the subdifferential of the $\ell_1/\ell_2$-block norm
over row $i$ takes the form
%
%
%e28 ###
\begin{eqnarray}
[\partial\| B \|_{\ell_{1} /\ell_{2}}]_i & = &
\cases{
\displaystyle \frac{{\beta}_i}{\|{\beta}_i\|_2}, &\quad\mbox{if }${\beta}_i
\neq\vec{0}$,\vspace*{2pt}
\cr
Z_i\mbox{ such that $\|Z_i\|_2 \leq1$}, &\quad\mbox{otherwise.}
}
\end{eqnarray}
%
% MJ This has already been defined, so it's not needed here.
% Note also that the definition has been augmented to allow zero rows.
%We also use the shorthand $\zfunc(\bBnew_i) = \Brow_i/\|\Brow_i\|_2$
%with
%an analogous definition for the matrix $\zfunc(\bBnew_{\Sset})$,
%assuming
%that no row of $\bBnew_{\Sset}$ is identically zero.
We define the \textit{empirical covariance matrix}
%
%
%e29 ###
\begin{eqnarray}
\widehat{\Sigma}& :=& \frac{1}{n} \bX ^T \bX  =
\frac{1}{n} \sum_{i=1}^n X_i X_i^T,
\end{eqnarray}
where $X_i$ is the $i$th row of $\bX $. (This definition is
natural under our standing assumption of zero mean for the variables
$X_i$; note, however, that our proofs extend readily to the case
of nonzero mean, in which case we would center the variables and use the
usual definition of the empirical covariance matrix.) We also make use
of the shorthand $\widehat{\Sigma}_{SS} = \frac{1}{n} \bX
_S^T
\bX _S$ and $\widehat{\Sigma}_{{S^c}S} = \frac{1}{n} \bX
_{S^c}^T
\bX _S$ as well as $\Pi_S=\bX _S(\widehat{\Sigma}_{SS
})^{-1} \bX _S^T$ to denote the projector on the range of
$\bX
_S$.

At the core of our constructive procedure is the following
convex-analytic result, which characterizes an optimal primal--dual
pair for which the primal solution $\widehat{B}$ correctly recovers the
support set $S$:
%l2
\begin{lems}
\label{LemUniqueOpt}
Suppose that there exists a primal--dual pair $(\widehat{B}, \widehat
{\bZ})$
that satisfies the conditions
%
%e30 ###
\begin{subequation}
\begin{eqnarray}
\label{EqnOnDual}
\widehat{\bZ}_S& = & \zeta(\widehat{B}_S), \\
\label{EqnOnPrimal}
-\lambda_n\widehat{\bZ}_{S} & = & \widehat{\Sigma}_{SS}
(\widehat{B}_{S}-B^*_{S})-\frac{1}{n} \bX _{S}^T
W,\\
\label{EqnOffDual}
\lambda_n\| \widehat{\bZ}_{{S^c}} \|_{\ell_{\infty} /\ell_{2}} &
:=& \bigg\| \widehat{\Sigma}_{{S^c}S} (\widehat{B}_{S}-B^*_{S})- \frac
{1}{n} \bX _{{S^c}}^T W \bigg\|_{\ell_{\infty} /\ell_{2}} <
\lambda_n,\\
\label{EqnOffPrimal}
\widehat{B}_{{S^c}} & = & 0.
\end{eqnarray}
\end{subequation}
Then $(\widehat{B}, \widehat{\bZ})$ is a primal--dual optimal solution
to the block-regularized problem, with $\widehat{S}(\widehat{B}) =
S$ by
construction. If $\widehat{\Sigma}_{SS} \succ0$, then $\widehat{B}$ is
the unique optimal primal solution.
\end{lems}

See Appendix~\ref{AppUniqueOpt} for the proof of this claim. Based on
Lemma~\ref{LemUniqueOpt}, we proceed to construct the required
primal--dual pair $(\widehat{B}, \widehat{\bZ})$ as follows. First,
we set
$\widehat{B}_{S^c}= 0$, so that condition~(\ref{EqnOffPrimal}) is
satisfied. Next, we specify the pair $(\widehat{B}_S,
\widehat{\bZ}_S)$ by solving the following restricted version of the
SOCP (\ref{EqnBlockRegProb}):
%
%
%e31 ###
\begin{eqnarray}
\label{EqnRestricted}
\widehat{B}_{S} = \arg\min_{ B_S\in\mathbb{R}^{s
\times
K}} \biggl\{ \frac{1}{2 n}  \bigg| \!  \bigg| \! \bigg| \bY-\bX
%{\scriptsize
%
\left[
\matrix{ B_S\vspace{2pt}\cr 0_{{S^c}}
}
\right]
%
%}
\bigg| \!  \bigg| \!  \bigg|_{{F}}^2 + \lambda_n\| B_S\|_{\ell_1 /\ell_2} \biggr\}.
\end{eqnarray}
Since $s< n$, the empirical covariance (sub)matrix
$\widehat{\Sigma}_{SS} = \frac{1}{n} \bX _{S}^T \bX
_{S}$
is strictly positive definite with probability one, which implies that
the restricted problem~(\ref{EqnRestricted}) is strictly convex and
therefore has a unique optimum $\widehat{B}_S$. We then choose
$\widehat{\bZ}_S$ to be the solution of
equation~(\ref{EqnOnPrimal}). Since any such matrix $\widehat{\bZ}_S$
is also a dual solution to the restricted SOCP~(\ref{EqnRestricted}),
it must be
an element of the subdifferential\vspace*{1pt}
$\partial\| \widehat{B}_S \|_{\ell_{1} /\ell_{2}}$.

It remains to show that this construction satisfies
conditions~(\ref{EqnOnDual}) and~(\ref{EqnOffDual}). In order to satisfy
condition~(\ref{EqnOnDual}), it suffices to show that no row of the
solution $\widehat{B}_S$ is identically zero. From
equation~(\ref{EqnOnPrimal}) and using the invertibility of the
empirical covariance matrix $\widehat{\Sigma}_{SS}$, we may solve as
follows:
%
%
%e32 ###
\begin{eqnarray}
\label{EqnDefnUvec}
(\widehat{B}_{S}-B^*_{S}) & = & (\widehat{\Sigma}_{S
S}
)^{-1} \biggl[ \frac{\bX _{S}^T W}{n} -
\lambda_n\widehat{\bZ}_{S} \biggr] = : \bU_S.
\end{eqnarray}
Note that for any row $i \in S$, by the triangle inequality, we have
\begin{eqnarray*}
\|\widehat{\beta}_i \|_2 & \geq& \|\beta^*_i\|_2 -
\| \bU_S \|_{\ell_{\infty} /\ell_{2}}.
\end{eqnarray*}
Therefore, in order to show that no row of $\widehat{B}_S$ is
identically zero, it suffices to show that the event
%
%
%e33 ###
\begin{eqnarray}
\label{EqnUcond}
\mathcal{E}(\bU_S) & :=& \bigl\{
\| \bU_S \|_{\ell_{\infty} /\ell_{2}} \leq \tfrac{1}{2}
b^*_{\min}
\bigr\}
\end{eqnarray}
occurs with high probability [recall from
equation~(\ref{EqnDefnBetamin}) that the parameter $b^*_{\min}$ measures
the minimum $\ell_2$-norm of any row of $B^*_S$]. We
establish this result in Section~\ref{SecUevent}.

Turning to condition~(\ref{EqnOffDual}), by substituting
expression~(\ref{EqnDefnUvec}) for the difference
$(\widehat{B}_{S}-B^*_{S})$ into
equation~(\ref{EqnOffDual}), we obtain a $(p- s) \times
K$ random matrix $V_{S^c}$, with rows indexed by
${S^c}$. For any index $j \in{S^c}$, the corresponding row
vector $V_j \in\mathbb{R}^K$ is given by
%
%
%e34 ###
\begin{eqnarray}
\label{EqnDefnVvec}
V_j & :=& \bX _{j}^T \biggl( [\Pi_{S} -I_n]
\frac{W}{n}- \lambda_n\frac{\bX _S}{n
}
(\widehat{\Sigma}_{SS})^{-1} \widehat{\bZ}_S\biggr).
\end{eqnarray}
In order for condition~(\ref{EqnOffDual}) to hold, it is necessary and
sufficient that the probability of the event
%
%
%e35 ###
\begin{eqnarray}
\label{EqnVcond}
\mathcal{E}(V_{S^c}) & :=&
\{\| V_{S^c} \|_{\ell_{\infty} /\ell_{2}} < \lambda_n
\}
\end{eqnarray}
converges to one as $n$ tends to infinity. Consequently, the
remainder (and bulk) of the proof is devoted to showing that the
probabilities $\mathbb{P}[\mathcal{E}(\bU_S)]$ and
$\mathbb{P}[\mathcal{E}(V_{S^c})]$ both converge to one under the
specified conditions.

%s3.2 ###
\subsection{Analysis of $\mathcal{E}(V_{S^c})$: Correct exclusion of nonsupport}
\label{SecVevent}

In this section, we prove the first claim of Theorem~\ref{ThmMain}(a),
namely that rows not in the support are always excluded. For
simplicity, in the following arguments, we drop the index ${S^c}$ and
write $V$ for $V_{S^c}$. In order to show that
$\| V \|_{\ell_{\infty} /\ell_{2}} < \lambda_n$ with probability
converging to one, we make use of the decomposition
$\frac{1}{\lambda_n} \| V \|_{\ell_{\infty} /\ell_{2}} \leq
\sum_{i=1}^3 T'_i$ where
%
%e36 ###
\begin{subequation}
\label{EqnDefnNewterm}
\begin{eqnarray}
T'_1 & :=& \frac{1}{\lambda_n} \| \EE [V | \bX _S] \|
_{\ell_{\infty} /\ell_{2}}, \\
T'_2 & :=& \frac{1}{\lambda_n}
\| \EE[V|\bX _S,W]-\EE[V|\bX _S ] \|_{\ell_{\infty} /\ell_{2}}, \\
T'_3 & :=& \frac{1}{\lambda_n}
\| V-\EE[V|\bX _S,W] \|_{\ell_{\infty} /\ell_{2}}.
\end{eqnarray}
\end{subequation}
We deal with each of these three terms in turn, showing that with high
probability under the specified scaling of $(n, p,
s)$, we have $T'_1 \leq(1-\mutinc)$, and $T'_2 =
o_p(1)$, and $T'_3 < \mutinc$, which suffices to show that
$\frac{1}{\lambda_n} \| V \|_{\ell_{\infty} /\ell_{2}} < 1$ with high
probability.

The following lemma is useful in the analysis:
%l3
\begin{lems}
\label{LemZapprox}
Define the matrix ${\Delta}\in\mathbb{R}^{s\times K}$ with
rows ${\Delta}_i :=\bU_i/\|\beta^*_i\|_2$. As long as
$\|{\Delta}_i\|_2 \leq1/2$ for all row indices $i \in S$,
we have
\begin{eqnarray*}
\| \widehat{\bZ}_S- \zeta(B^*_S) \|_{\ell_{\infty} /\ell_{2}} &
\leq& 4
\| {\Delta} \|_{\ell_{\infty} /\ell_{2}}.
\end{eqnarray*}
%
%Hence $\Genblnorm{\bDelta}{\infty}{2} = o_p(1)$ implies that
%$\Genblnorm{\Dualmat_\Sset- \zfunc(\bBstar_\Sset)}{\infty}{2} =
%o_p(1)$.
%
\end{lems}

See Appendix~\ref{AppLemZapprox} for the proof of this
claim.

%%%%%%%%%%%%%%%%%%%

%s3.2.1 ###
\subsubsection{Analysis of $T'_1$}

Note that by definition of the regression
model~(\ref{EqnGenObsModel}), we have the conditional independence
relations
\[
W\indep\bX _{{S^c}} \vert\bX _S,\qquad
\widehat{\bZ}_S\indep\bX _{{S^c}} \vert \bX _S\quad \mbox{and}\quad
\widehat{\bZ}_S\indep\bX _{{S^c}} \vert \{\bX _S,W\}.
\]
Using the two first conditional independencies, we have
\begin{eqnarray*}
\EE[V\vert\bX _S] & = & \EE[\bX _{S^c}^T | \bX _S
]
\biggl( [\Pi_{S} -I_n] \frac{\EE[W| \bX _S]}{n}-
\lambda_n\frac{\bX _S}{n} (\widehat{\Sigma}_{S
S})^{-1} \EE[\widehat{\bZ}_S| \bX _S] \biggr).
\end{eqnarray*}
Since $\EE[W| \bX _S]=0$, the first term vanishes, and using
$\EE[\bX _{S^c}^T | \bX _S]= \Sigma_{{S^c}S}
{\Sigma}_{SS}^{-1}\bX _S^T$, we obtain
%
%
%e37 ###
\begin{equation}
\EE[V\vert\bX _S]=\lambda_n\Sigma_{{S^c}S}
{\Sigma}_{SS}^{-1} \EE[\widehat{\bZ}_S|\bX _S].
\end{equation}
Using the matrix-norm inequality~(\ref{EqnSubordinate}) from
Appendix~\ref{AppLemGuillaume} and then Jensen's inequality yields
%
%
%e38 ###
\begin{eqnarray}\label{EqnNewtermoneBound}
T'_1 & = & \| \Sigma_{{S^c}S} {\Sigma}_{SS}^{-1} \EE[Z_S
\vert X_S]\|_{\ell_\infty/\ell_2} \nonumber\\
& \leq&  | \! | \! | \Sigma_{{S^c}S} {\Sigma}_{SS}^{-1}
 | \!  | \!  |_{{\infty }}
\EE[ \| Z_S\|_{\ell_\infty/\ell_2} \vert X_S]
\\
& \leq& (1-\mutinc).\nonumber
\end{eqnarray}

%s3.2.2 ###
\subsubsection{Analysis of $T'_2$}

Appealing to the conditional independence relationship $\widehat{\bZ}_S
\indep\bX _{{S^c}} | \{\bX _S,W\}$, we have
\begin{eqnarray*}
&&\EE[V|\bX _S,W]
\\
&&\qquad= \EE[\bX _{S^c}^T
|
\bX _S,W] \biggl( [\Pi_{S} -I_n]
\frac{W}{n}- \lambda_n\frac{\bX _S}{n
}
(\widehat{\Sigma}_{SS})^{-1} \EE[\widehat{\bZ}_S| \bX _S
,W]
\biggr).
\end{eqnarray*}
Observe that $\EE[\widehat{\bZ}_S| \bX _S,W]=\widehat{\bZ}
_S$
because $(\bX _S,W)$ uniquely specifies $\widehat{B}_S$
through the convex program~(\ref{EqnRestricted}), and the triple
$(\bX _S,W,\widehat{B}_S)$ defines $\widehat{\bZ}_S$ through
equation~(\ref{EqnOnPrimal}). Moreover, the noise term disappears
because the
kernel of the orthogonal projection matrix
$(I_n-\Pi_{S})$ is the same as the range space of $\bX
_S$,
and
\begin{eqnarray*}
\EE[\bX _{S^c}^T \vert \bX _S,W] [\Pi_{S}
-I_n] & = & \EE[\bX _{S^c}^T \vert
\bX _S] [\Pi_{S} -I_n] \\
& = & \Sigma_{{S^c}S} {\Sigma}_{SS}^{-1}\bX _S^T [\Pi_{S}
-I_n] = 0.
\end{eqnarray*}
We have thus shown that $\EE[V|\bX _S,W] = -
\frac{\lambda_n}{n} \Sigma_{{S^c}S}
{\Sigma}_{SS}^{-1}\widehat{\bZ}_S$, so that we can conclude that
%
%
%e39 ###
\begin{eqnarray}
T'_2 & \leq&  | \! | \! | \Sigma_{{S^c}S} (\Sigma _{S
S})^{-1}  | \!  | \!  |_{{\infty}}
\|\widehat{\bZ}_S-\EE[\widehat{\bZ}_S|\bX _S]\|_{\ell_\infty/\ell_2}
\nonumber\\
\label{EqnNewtermtwoBound}
& \leq& (1 - \mutinc) \EE[
\| \widehat{\bZ}_S-\bZ^*_S \|_{\ell_{\infty} /\ell_{2}} ] + (1 -
\mutinc) \| \widehat{\bZ}_S-\bZ^*_S \|_{\ell_{\infty} /\ell_{2}}
\\
& \leq& (1- \mutinc) 4 \{ \EE[
\| {\Delta} \|_{\ell_{\infty} /\ell_{2}} ] + \| {\Delta} \|_{\ell_{\infty} /\ell_{2}}
\}, \nonumber
\end{eqnarray}
where the final inequality uses Lemma~\ref{LemZapprox}. Under the
assumptions of Theorem~\ref{ThmMain}, this final term is of order
$o_p(1)$, as will be shown in Section~\ref{SecUevent}.

%s3.2.3 ###
\subsubsection{Analysis of $T'_3$}

This third term requires a little more care. We begin by noting that
conditionally on $\bX _S$ and $W$, each vector $V_j \in
\mathbb{R}^K$ is normally distributed. Since $\operatorname
{Cov}(\bX ^{(j)}
|
\bX _S, W) = (\Sigma_{{S^c}\mid S})_{jj}
I_n$,
we have
\begin{eqnarray*}
\operatorname{Cov}(V_j \vert \bX _S, W) & = & {M}_n
(\Sigma_{{S^c}\mid S})_{jj},
\end{eqnarray*}
where the $K\times K$ random matrix ${M}_n=
{M}_n(\bX _S, W)$ is given by
%
%
%e40 ###
\begin{eqnarray}
\label{EqnDefnMat}
{M}_n& :=& \frac{\lambda_n^2}{n}
{\widehat{\bZ}_S}^T (\widehat{\Sigma}_{SS})^{-1} \widehat{\bZ}_S+
\frac{1}{n^2} W^T (\Pi_{S}- I_n) W.
\end{eqnarray}

We begin by noting that by its definition~(\ref{EqnOnDual}), the
candidate dual matrix $\widehat{\bZ}_S$ is a function only of $W$
and $\bX _S$. Therefore, conditioned on the pair $(W,
\bX _S)$, the matrix ${M}_n$ is fixed, and we have
%
%
%e41 ###
\begin{eqnarray}
\label{EqnKeyRelation}
( \|V_j-\EE[V_j \vert \bX _S, W]\|^2_2
| W, \bX _S ) & \eqdis& (\Sigma_{S^c
S^c | S} )_{jj} \xi_j^T {M}_n \xi_j,
\end{eqnarray}
where $\xi_j \sim N(\vec{0}_K, I_K)$. By definition of
$\rho_{u }(\Sigma_{S^c S^c | S})=\max_{j}
(\Sigma_{S^c S^c | S})_{jj}$, we have $(\Sigma_{S^c
S^c | S})_{jj} \leq\rho_{u }(\Sigma_{S^c S^c |
S}) \leq C_{\max}$ and
\begin{eqnarray*}
\max_{j \in{S^c}} (\Sigma_{S^c S^c | S
})_{jj} \xi_j^T {M}_n\xi_j & \leq&
\rho_{u }(\Sigma_{S^c S^c | S})
 | \!  | \! | {M}_n  | \!  | \!  |_{{2}}{} \max
_{j \in{S^c}} \|\xi_j\|_2^2,
\end{eqnarray*}
where $ | \!  | \! | {M}_n  | \!  | \!
|_{{2}}{}$ is the spectral norm.

We now state a result that provides control on this spectral norm, in
particular showing that the rescaled random matrix\vspace*{-3pt}
$\frac{n}{\lambda_n^2}{M}_n$
concentrates around the deterministic matrix ${M}^* :=
{\bZ^*_S}^T (\Sigma_{SS})^{-1} \bZ^*_S$. This
concentration establishes the link to the sparsity-overlap
function~(\ref{EqnDefnGenKeyFunc}), which is given by the spectral
norm $ | \!  | \! | {M}^*  | \!  | \!
|_{{2}}$. For any $\delta\in(0,1)$, define the
event
%
%
%e42 ###
\begin{equation}
\label{EqnDefnTail}
\hspace*{30pt}\mathcal{T}(\delta) := \biggl\{ \frac{\lambda_n^2
\psi(B^*)+ \sigma^2}{n} (1- \delta) \leq
 | \!  | \! | {M}_n  | \!  | \!  |_{{2}}{} \leq
\frac{\lambda_n^2
\psi(B^*)+ \sigma^2}{n} (1+\delta) \biggr\}.
\end{equation}
Moreover, recall the definition of $\Delta$ from
Lemma~\ref{LemZapprox}. The following result provides sufficient
conditions for the event $\mathcal{T}(\delta)$ to hold with high
probability.
%l4
\begin{lems}
\label{LemTailBound}
Suppose that $\frac{s}{n}=o(1)$ and
$\| \Delta \|_{\ell_{\infty} /\ell_{2}} = o(1)$. Then for any
$\delta\in
(0,1)$, there is some $c_1 = c_1(\delta) > 0$ such that
$\mathbb{P}[\mathcal{T}(\delta)^c] \leq c_1 \exp(- c_0 K
\log
s)
\rightarrow0$.
\end{lems}

See Appendix~\ref{AppLemTailBound} for the proof of this
lemma.

Given the assumptions of Theorem~\ref{ThmMain} and the
bound~(\ref{EqnKeyUBound}), we observe that the hypotheses of
Lemma~\ref{LemTailBound} are satisfied, and we can now complete the\vadjust{\goodbreak}
proof. For any fixed but arbitrarily small $\delta> 0$, we have
\begin{eqnarray*}
\mathbb{P}[ T'_3 \geq\mutinc] & \leq& \mathbb{P}[T'_3 \geq
\mutinc \vert \mathcal{T}(\delta)] + \mathbb{P}[\mathcal{T}(\delta)^c].
\end{eqnarray*}
Since $\mathbb{P}[\mathcal{T}(\delta)^c] \rightarrow0$ from
Lemma~\ref{LemTailBound}, it suffices to deal with the first term.
Conditioning on the event $\mathcal{T}(\delta)$, we have
\begin{eqnarray*}
\mathbb{P}[T'_3 \geq\mutinc \vert \mathcal{T}(\delta)] & \leq&
\mathbb{P}\biggl[\max_{j \in{S^c}} \|\xi_j\|_2^2 \geq
\frac{\mutinc^2}{\rho_{u }(\Sigma_{S^c S^c | S})}
\frac{n}{( \psi(B^*) + \sfrac{\sigma^2}{\lambda
_n^2} )
(1+\delta)} \biggr].
\end{eqnarray*}
Now define the quantity
\[
t^*(n, B^*) := \frac{1}{2}
\frac{\mutinc^2}{\rho_{u}(\Sigma_{S^c S^c | S})}
\frac{n}{( \psi(B^*) + \sfrac{\sigma^2}{\lambda
_n^2} )
(1+\delta)},
\]
and note that $t^* \rightarrow+\infty$ under the specified scaling of
$(n, p, s)$. By applying Lemma~\ref{LemChiMax} from
Appendix~\ref{AppChiMax} on large deviations for $\chi^2$-variates
with $t = t^*(n, B^*)$, we obtain
%
%
%e43 ###
\begin{eqnarray}\label{EqnFinalBound}
\mathbb{P}[T'_3 \geq\mutinc \vert \mathcal{T}(\delta)] & \leq&
(p- s) \exp\Biggl(-t^* \Biggl[1- 2
\sqrt{\frac{K}{t^*}} \Biggr] \Biggr) \nonumber\\[-8pt]\\[-8pt]
& \leq& (p- s) \exp\bigl( -t^* (1- \delta) \bigr)\nonumber
\end{eqnarray}
for $(n, p, s)$ sufficiently large. Now denoting
$\theta_u :={\rho_{u}(\Sigma_{S^c S^c | S
})}/{\mutinc^2}$, we have, by assumption, that $n\geq2 (1+\nu)
\theta_u \psi(B^*) \log(p- s)$. Given that
$\lambda_n^2=\frac{f(p) \log(p)}{n}$, we have
$\frac{\sigma^2}{\lambda_n^2} \log(p-s) \leq\sigma^2
\frac{n}{f( p)}=o(n)$ so that for any $\epsilonn>0$, we have
\begin{eqnarray*}
n \geq\frac{1+\nu}{1+\epsilonn} \biggl(2 \theta_u\psi(B^*) \log(p- s)+\frac{2\sigma^2}{\lambda_n^2} \log(p- s) \biggr)
\end{eqnarray*}
once $n$ is sufficiently large. This inequality implies that $
(1-\delta) t^*(n, B^*) \geq
\frac{(1+\nu)(1-\delta)}{(1+\epsilonn)(1+\delta)} \log(p-
s)$. Thus for $\delta$ and $\epsilonn$ sufficiently small, the
bound~(\ref{EqnFinalBound}) tends to zero at rate $\mathcal{O}(\exp
(-\nu/2
\log(p- s)))$ which establishes the claim.
% deleted equations
% [ 2 \keyfunc(\bBstar) \log(\pdim- \spindex) ].
%This establishes or claim, because, given that $\regpar_n^2=\frac{f(

%%%%%%%%%%%%%%%%%%%%%%%%%%%%%%%%%%%%%%%%%%%%%%%%%%%%%%%%%%%%%%%%%%%%%%

%s3.3 ###
\subsection{Analysis of $\mathcal{E}(\bU_S)$: Correct inclusion of supporting covariates}\label{SecUevent}

This section is devoted to the analysis of the event
$\mathcal{E}(\bU_S)$ from equation~(\ref{EqnUcond}), and in particular
showing that its probability converges to one under the specified
scaling. This allows us to establish the $\ell_2/\ell_\infty$ bound
in Theorem~\ref{ThmMain}(a), as well as the correct support recovery
claim in part (b).

If we define the noise matrix $\widetilde{W}:=\frac{1}{\sqrt{n}}
(\widehat{\Sigma}_{SS})^{-\sfrac{1}{2}} \bX _S^T W$, then we
have
\begin{eqnarray*}
\bU_S& = & \widehat{\Sigma}_{SS}^{-\sfrac{1}{2}}
\frac{\widetilde{W}}{\sqrt{n}}- \lambda_n(\widehat{\Sigma}_{SS
})^{-1}\widehat{\bZ}_S.
\end{eqnarray*}
Using this representation and the triangle inequality, we obtain
\begin{eqnarray*}
\| \bU_S \|_{\ell_{\infty} /\ell_{2}} & \leq&
\bigg\| (\widehat{\Sigma} _{S S})^{-\sfrac{1}{2}} \frac{\widetilde{W}}{\sqrt{n}} \bigg\|_{\ell_{\infty} /\ell_{2}} +
\lambda_n\| (\widehat{\Sigma}_{S S})^{-1}\widehat{\bZ}_S \|_{\ell_{\infty} /\ell_{2}} \\
& \leq&
\mathop{\underbrace{{\bigg\| (\widehat{\Sigma}_{SS})^{-\sfrac{1}{2}}\frac{\widetilde{W}}{\sqrt{n}} \bigg\|_{\ell_{\infty} /\ell_{2}}}_{}}}_{T_1} +
\mathop{\underbrace{{\lambda_n | \! | \! | (\widehat{\Sigma}_{SS})^{-1}  | \!| \!  |_{\infty}}_{}}}_{T_2},
\end{eqnarray*}
where the form of $T_2$ in the second line uses a standard matrix
norm bound [see equation~(\ref{EqnSubordinate}) in
Appendix~\ref{AppLemGuillaume}], and the fact that
$\| \widehat{\bZ}_S \|_{\ell_{\infty} /\ell_{2}} \leq1$.

Using the triangle inequality, we bound $T_2$ as follows:
\begin{eqnarray*}
T_2 & \leq& \lambda_n \{  | \! | \! | (\Sigma _{S
S})^{-1}  | \!  | \!  |_{{\infty}} +  | \! | \!
 | (\widehat{\Sigma}_{SS })^{-1}-(\Sigma_{SS})^{-1}  | \!
 | \!  |_{{\infty}} \} \\
& \leq& \lambda_n \bigl\{ D_{\max}+ \sqrt{s}
 | \! | \! | (\widehat{\Sigma}_{SS})^{-1}-(\Sigma_{S
S})^{-1}  | \!  | \!  |_{{2}} \bigr\} \\
& \leq& \lambda_n \bigl\{ D_{\max}+ \sqrt{s}
 | \! | \! | (\Sigma_{SS})^{-1}  | \!  | \!  |_{{2}}
 | \! | \! | (\widetilde{\bX }_S^T \widetilde
{\bX }_S/n)^{-1}-I_{s}  | \!  | \!  |_{{2}} \bigr\} \\
& \leq& \lambda_n \biggl\{ D_{\max}+
\frac{\sqrt{s}}{C_{\min}}  | \! | \! | (\widetilde
{\bX }_S^T \widetilde{\bX }_S/n)^{-1}-I_{s}  | \!
 | \!  |_{{2}} \biggr\},
\end{eqnarray*}
which defines $\widetilde{\bX }_S$ as a random matrix with
i.i.d.\ standard
Gaussian entries. From concentration results in random matrix theory
(see Appendix~\ref{AppRandMat}), for \mbox{$s/n\rightarrow0$},
we have $ | \! | \! | (\widetilde{\bX }_S^T
\widetilde{\bX }_S/n)^{-1}-I_{s}  | \!  | \!
|_{{2}} \leq6
\sqrt{\frac{s}{n}}$ with probability $1-2
\exp(-s/2 )-\exp(-\Theta(n))$. Overall, we conclude that
\begin{eqnarray*}
T_2 & \leq& \lambda_n \Biggl\{ D_{\max}+ \frac{6}{C_{\min}}
\sqrt{\frac{s^2}{n}} \Biggr\}
\end{eqnarray*}
with probability $1-2\exp(-s/2)-\exp(-\Theta(n))$.

Turning now to $T_1$, let us introduce the notation
$\operatorname{vec}(A)$ to denote the vectorized version of a matrix $A$,
obtained by stacking all of its rows into a single vector.
Conditioning on $\bX _S$, we have $(\operatorname{vec}(\widetilde{W})
|
\bX _S) \sim N(\vec{0}_{s\times K}, I_s
\otimes I_K)$. Combined with the definition of the block
$\ell_\infty/\ell_2$ norm, we obtain
\begin{eqnarray*}
T_1 & = & \max_{i \in S} \bigg\|e_i^T (\widehat{\Sigma}_{S S})^{-\sfrac{1}{2}} \frac{\widetilde{W}}{\sqrt{n}} \bigg\|_2
\leq
 | \! | \! | (\widehat{\Sigma}_{SS})^{-1}  | \!
| \!  |_{{2}}^{1/2}
\biggl[\frac{1}{n}\max_{i \in S} \zeta_i^2 \biggr]^{1/2},
\end{eqnarray*}
where the variates $\{\zeta_i^2\}$ are an i.i.d. sequence of
$\chi^2$-variates with $K$ degrees of freedom. Using the tail
bound in Lemma~\ref{LemChiMax} (see Appendix~\ref{AppChiMax}) with \mbox{$t
= 2 K\log s> K$}, we have
\begin{eqnarray*}
\mathbb{P}\biggl[ \frac{1}{n}\max_{i \in S} \zeta_i^2 \geq
\frac{4 K\log s}{n} \biggr] & \leq& \exp\biggl(- 2
K\log s\bigl(1- 2 (2 \log s)^{-1/2} \bigr) \biggr)
\rightarrow0.
\end{eqnarray*}
Define the event $\mathcal{T}:=\{  | \! | \! | (\widehat
{\Sigma}_{S S})^{-1}  | \!  | \!  |_{{2}} \leq\frac
{2}{C_{\min}} \}$; the bound
$\mathbb{P}[\mathcal{T}] \geq1- \exp(-\Theta(n))$ then follows
from known
concentration results in random matrix theory (see
Appendix~\ref{AppRandMat}). Thus, we obtain
%
%
%e44 ###
\begin{eqnarray}\label{EqnNoiseTerm}
\mathbb{P}\Biggl[T_1 \geq\sqrt{
\frac{8 K\log s}{C_{\min}
n}} \Biggr] & \leq& \mathbb{P}\Biggl[ T_1 \geq\sqrt{\frac{8
K\log s}{C_{\min}n}} \Big\vert \mathcal{T}\Biggr] +
\mathbb{P}[\mathcal{T}^c] \nonumber\\
& \leq& \mathbb{P}\biggl[\frac{1}{n} \max_{i \in S} \zeta_i^2
\geq\frac{4 K\log s}{n} \biggr] + \exp
\Biggl\{ -n\Biggl(\frac{1}{2}-\sqrt{\frac{s}{n}}
\Biggr) \Biggr\} \\
& = & \mathcal{O}(\exp(- c_0 K\log s) )
\rightarrow0,\nonumber
\end{eqnarray}
where $c_0 > 0$ is a universal constant. Combining the pieces, we
conclude with probability $1-\exp(- c_0 K\log s)$, we
have
\begin{eqnarray*}
\| \bU_S \|_{\ell_{\infty} /\ell_{2}} & \leq&
\frac{1}{b^*_{\min}} [T_1 + T_2 ]
 \leq \Biggl[ \sqrt{\frac{8 K\log
s}{C_{\min}n}} + \lambda_n \Biggl(D_{\max}+
\frac{6}{C_{\min}} \sqrt{\frac{s^2}{n}} \Biggr)
\Biggr] \\
& = & \myerror(n, s, \lambda_n),
\end{eqnarray*}
which establishes the bound~(\ref{EqnMyerror}) from
Theorem~\ref{ThmMain}(a).

Moreover, under the assumptions of Theorem~\ref{ThmMain}(b), we
can conclude that
%
%
%e45 ###
\begin{equation}
\label{EqnKeyUBound}
\frac{\| \bU_S \|_{\ell_{\infty} /\ell_{2}}}{b^*_{\min}} \leq
\frac{\myerror(n, s, \lambda_n)}{b^*_{\min}} =
o(1),
\end{equation}
with probability greater than $1 - \Theta(\exp(-c_0 K\log
s)) \rightarrow1$. Consequently, the conditions of
Theorem~\ref{ThmMain}(b) are sufficient to ensure that the event
$\mathcal{E}(\bU_S)$ holds with high probability as claimed.

\begin{remark*}
As we noted following the statement of
Theorem~\ref{ThmMain}, the fact that the claims hold with probability
converging to one only if $s\rightarrow+\infty$ might appear
counter-intuitive and does not allow the result to cover problems with
fixed sizes $s$ of the row support. Here we discuss how this
condition can be weakened. Note that our assumptions imply that
$p-s\rightarrow\infty$ and that
$\frac{s}{n}=o(1)$. Consequently, for any $a>0$, we have
$\frac{\log s}{n^a}=\frac{\log s}{s^a}
\frac{s^a}{n^a}=o(1)$, so that we may use a slightly
weaker bound on $T_1$ in equation~(\ref{EqnNoiseTerm}). Indeed,
with the same notation as in that equation, we have
\begin{eqnarray*}
&&\mathbb{P}\Biggl[ T_1 \geq\sqrt{\frac{4( K+\log
s+n^a)}{C_{\min}n}} \Big\vert \mathcal{T}\Biggr]
\\
&&\qquad\leq \mathbb{P}\biggl[ \frac{1}{n} \max_{i \in S} \zeta_i^2 \geq
\frac{2}{n}( K+\log s+n^a) \big\vert
\mathcal{T}
\biggr] \\
&&\qquad \leq \exp\Biggl\{- n^a \Biggl( 1-2
\biggl(1+\frac{\log s}{n^a} \biggr)
\sqrt{\frac{K}{K+n^a}}  \Biggr) \Biggr\}
\rightarrow0,
\end{eqnarray*}
where the last inequality is obtained by setting $t=K+\log
s+n^a$ in Lemma~\ref{LemChiMax} of
Appendix~\ref{AppChiMax}.
\end{remark*}

%s4 ###
\section{\texorpdfstring{Proof of Theorem~\protect\ref{ThmNecessary}}{Proof of Theorem~2}}
\label{SecProofThmNecessary}

In this section, we prove the necessary conditions stated in
Theorem~\ref{ThmNecessary}. We begin by noting that we may assume
without loss of generality that $s< n$, since it is
otherwise impossible to recover the support (even in the absence of
noise). In order to develop some intuition for the argument to
follow, recall the definition~(\ref{EqnVcond}) of the event
$\mathcal{E}(V_{S^c})$. The proof of Theorem~\ref{ThmNecessary} is based
on the fact that if $\mathcal{E}(V_{S^c})$ does not hold, then no
solution of the multivariate group Lasso has the correct row support.

Again, to lighten notation, we write $V$ for the quantity
$V_{S^c}$. Recall the definitions~(\ref{EqnDefnNewterm}) of the
quantities $T'_i$ for $i=1$, $2$ and $3$. By the triangle
inequality, we have
%
%
%e46 ###
\begin{eqnarray}
\label{EqnTriangleTwo}
\frac{1}{\lambda_n} \| V \|_{\ell_{\infty} /\ell_{2}} & \geq&
T'_3 - T'_2 - T'_1.
\end{eqnarray}
From our earlier argument [see equation~(\ref{EqnNewtermoneBound})], we
know that $T'_1 \leq(1-\mutinc)$. From the
bound~(\ref{EqnNewtermtwoBound}), in order to show that $T'_2 =
o(1)$ with high probability, it suffices to show that
$\| \widehat{\bZ}_S- \bZ^*_S \|_{\ell_{\infty} /\ell_{2}}
= o(1)$.
We reason by contradiction and assume that in the regime considered in
Theorem~\ref{ThmNecessary}, there is a solution of the multivariate
group Lasso which satisfies $\| \widehat{B} - B^* \|_{\ell_{\infty} /\ell_{2}} = o(b^*_{\min})$ with high probability. Note that
this condition implies that $\max_{i \in S}
\frac{\|\widehat{B}_i - B^*_i \|_2}{\|B^*_i\|_2} = o(1)$, so
that we may apply Lemma~\ref{LemZapprox} to conclude that
$\| \widehat{\bZ}_S- \bZ^*_S \|_{\ell_{\infty} /\ell_{2}} = o(1)$
as well.
Consequently, we conclude that $T'_2 = o(1)$.

Considering the
decomposition~(\ref{EqnTriangleTwo}), we obtain that
%
%
%e47 ###
\begin{eqnarray}
\label{EqnKeyLower}
T'_3-T'_2-T'_1 & = & \frac{1}{\lambda_n}
\| V-\EE[V|\bX _S,W] \|_{\ell_{\infty} /\ell_{2}} - (1-
\mutinc) - o(1).
\end{eqnarray}
Therefore, it suffices to prove that $T'_3 > 2-\mutinc$. The remainder
of the proof is devoted to establishing this claim.

In order to analyze $T'_3$, let us recall the notation
$\widetilde{V}_j=V_j-\EE[V_j|X_{S},W]$, where for each $j \in{S^c}$, the
quantity $V_j \in\mathbb{R}^{K}$ denotes the $j$th row of the
matrix $V$. As shown earlier in Section~\ref{SecVevent}, we
can write
\begin{eqnarray*}
( \|\widetilde{V}_j\|_2^2 | W, \bX _S ) & \eqdis&
\Sigma_{jj | S} \xi_j^T {M}_n \xi_j,
\end{eqnarray*}
where for each $j \in{S^c}$, the random vector $\xi_j \sim N(0,
I_{K})$. The random vectors $(\xi_j, j \in{S^c})$ are
not i.i.d. in general, since for each pair $i, j \in{S^c}$, we
have $\operatorname{cov}(\xi_i, \xi_j) = \frac
{\Sigma_{ij \mid S}}{\sqrt{\ensuremath{\Sigma
}_{ii \mid S}\Sigma_{jj\mid S}}} I_{K}$.

The next part of the proof is devoted to analyzing the behavior of the
random variable
%
%
%e48 ###
\begin{eqnarray}
\label{EqnDefnVmax}
V_{\max}& :=& \max_{j \in{S^c}} \|\widetilde{V}_j\|_2 = \max_{j
\in
{S^c}} \sqrt{ \Sigma_{jj \mid S} \xi^T_j {M}_n
\xi_j },
\end{eqnarray}
with our goal in particular being to show that
$\frac{V_{\max}}{\lambda_n} \geq2 - \mutinc$ with high
probability. In order to lower bound the random variable $V_{\max}$, our
first step is to show that it is sharply concentrated around its
expectation.
%l5
\begin{lems}
\label{LemVmaxLip}
For any $\delta> 0$, we have
%
%
%e49 ###
\begin{equation}
\label{EqnVmaxLip}
\hspace*{15pt}\PP\bigl[ |V_{\max}- \mathbb{E}[V_{\max}]| \geq\delta | X_{S}, W
\bigr] \leq4 \exp\biggl\{ -\frac{1}{2} \frac{\delta^2}{
\rho_{u }({\Sigma}_{{S^c}{S^c} \mid S})
 | \!  | \! | {M}_n  | \!  | \!  |_{{2}}} \biggr\},
\end{equation}
where $\rho_{u }(\Sigma_{{S^c}{S^c} \mid S
}) = \max_{j \in
{S^c}} \Sigma_{jj \mid S}$.
\end{lems}
\begin{pf}
By standard Gaussian concentration theorems [e.g., Theorem 3.8
of~\citet{Massart03}], if $X$ has a standard Gaussian measure on
$\mathbb{R}^{m}$ and $f$ is a Lipschitz function with Lipschitz
constant $L$,
then
%
%
%e50 ###
\begin{equation}
\label{EqnLipschitz}
\PP\bigl[ |\EE[f(X)]-f(X) | \geq x\bigr] \leq4 \exp\bigl(- x^2/(2 L^2)
\bigr).
\end{equation}
In order to exploit this result in application to $V_{\max}$, we consider
the function $f\dvtx  \mathbb{R}^{(p- s) \times K}
\rightarrow\mathbb{R}$ defined by
\[
f(\xi_j, j \in{S^c}) :=\max_{j \in{S^c}} \sqrt{\Sigma_{jj
\mid S}} \big\|\sqrt{{M}_n} \xi_j \big\|_2,
\]
which is equal to $V_{\max}$ by construction. Let $u = (u_j, j \in
{S^c})$ and $v = (v_j, j \in{S^c})$ be two collections of vectors.
We have
\begin{eqnarray*}
|f(u)-f(v)| & = & \max_{j \in{S^c}} \sqrt{\Sigma_{jj \mid S}}
\big\|\sqrt{{M}_n} u_j \big\|_2 - \max_{k \in{S^c}} \sqrt{\Sigma_{kk \vert S}} \big\|\sqrt{{M}_n} v_k \big\|_2
\\
& \leq& \max_{j \in{S^c}} \sqrt{\Sigma_{jj \mid S}}
\big\|\sqrt{{M}_n} (u_j - v_j)\big\|_2 \\
& \leq& \sqrt{\rho_{u }(\Sigma_{{S^c}{S^c} \vert
S})}
\sqrt{ | \!  | \! | {M}_n  | \!  | \!
|_{{2}}} \|u-v\|_2.
\end{eqnarray*}
We may therefore apply the bound~(\ref{EqnLipschitz}) with $L^2
=  | \!  | \! | M_n  | \!  | \!  |_{{2}} \rho
_{u }(\Sigma_{{S^c}{S^c}
\mid
S})$ to obtain the claim.
\end{pf}

The second key ingredient in our proof is a lower bound on the
expected value of $V_{\max}$:
%l6
\begin{lems}
\label{LemVmaxExp}
For any fixed $\delta'> 0$, with probability $1-o(1)$ as $(p-
s) \rightarrow+\infty$, we have
%
%
%e51 ###
\begin{equation}
\label{EqnVmaxExp}
\EE[ V_{\max} \vert X_S,W] \geq\sqrt
{ | \!  | \! | {M}_n  | \!  | \!  |_{{2}}}
\sqrt{2 (1-\delta')
\rho_{\ell}({\Sigma}_{S^c S^c | S})
\log(p-s)}.
\end{equation}
\end{lems}
\begin{pf}
We may diagonalize ${M}_n$, writing ${M}_n= U^T D U$,
where $U \in\mathbb{R}^{K\times K}$ is orthogonal, and $D =
\operatorname{diag}\{d_1, \ldots, d_K\}$ is diagonal with $d_1 =
 | \!  | \! | {M}_n  | \!  | \!  |_{{2}}$.
Since the distribution of the
$K$-dimensional normal vector $\xi_j \sim N(0, I)$ remains
invariant under orthogonal transformations, for each $j \in{S^c}$, we
can write
\begin{eqnarray*}
\sqrt{\Sigma_{jj \mid S} \xi_j^T {M}_n\xi_j} &
\stackrel{d}{=} & \sqrt{\Sigma_{jj \mid S} \eta_j^T
D \eta_j} \geq \sqrt{ \Sigma_{jj \mid S}
 | \!  | \! | {M}_n  | \!  | \!  |_{{2}}} |\eta_{j,1}|,
\end{eqnarray*}
where $\eta_{j,1} \sim N(0, \Sigma_{jj \vert S})$.
Overall, we have
\begin{eqnarray*}
\EE[ V_{\max} \vert X_S,W] & = & \mathbb{E}\Bigl[ \max_{j \in{S^c}}
\sqrt{\Sigma_{jj | S} \xi_j^T {M}_n\xi_j}
| X_S, W\Bigr]
 \geq \sqrt{  | \!  | \! | {M}_n  | \!  | \! |_{{2}}} \mathbb{E}\Bigl[\max_{j \in{S^c}}|\eta_{j,1}|\Bigr],
\end{eqnarray*}
where the vector $\eta= (\eta_{j,1}, j \in{S^c})$ is
zero-mean Gaussian with covariance $\Sigma_{{S^c}{S^c}
|
S}$.

Our next step is to lower bound the expectation $\mathbb{E}[\max_{j
\in{S^c}} |\eta_{j,1}|]$ by a Gaussian comparison argument, in
particular exploiting the Sudakov--Fernique
inequality~[\citet{LedTal91}]. Let $\widetilde{\eta}\in
\mathbb{R}^{p
- s}$ be a Gaussian random vector with i.i.d. $N(0,1)$ entries.
By the definition~(\ref{EqnRhoLower}) of $\rho_{\ell}(\cdot)$, we have
\begin{eqnarray*}
\EE[(\eta_i - \eta_j)^2] & = & \Sigma_{ii \mid S}
-2 \Sigma_{ij \mid S} + \Sigma_{jj \mid S} \\
& \geq& \rho_{\ell}(\Sigma_{{S^c}{S^c}\mid S
} ) \EE[
(\widetilde{\eta}_{j}-\widetilde{\eta}_{i})^2]\qquad \mbox{for all $i,j$.}
\end{eqnarray*}
Consequently, the Sudakov--Fernique inequality implies that
\[
\EE\Bigl[ \max_{j \in{S^c}} | \eta_{j} | \Bigr]
\geq
\sqrt{\rho_{\ell}(\Sigma_{{S^c}{S^c}\mid S
} )} \EE
\Bigl[ \max_{j \in{S^c}} |\widetilde{\eta}_j | \Bigr].
\]
From standard results on Gaussian extrema~[\citet{LedTal91}], for any
fixed $\delta'\in(0,1)$, we have $\EE[ \max_{j \in
{S^c}} |\widetilde{\eta}_j | ] \geq\sqrt{2 (1-\delta')
\log(p- s)}$ once \mbox{$(p- s)$} is sufficiently large,
which completes the proof.
\end{pf}

It remains to show that the random matrix $ | \!  | \! |
{M}_n  | \!  | \!  |_{{2}}$
previously defined~(\ref{EqnDefnMat}) is suitably concentrated. Our
approach is to show that unless the hypotheses of
Lemma~\ref{LemTailBound}---namely, $s/n= o(1)$ and
$\| \widehat{\bZ}-Z^* \|_{\ell_{\infty} /\ell_{2}} = o(1)$---are
both satisfied,
then the multivariate group Lasso fails. We have shown previously that
the latter
condition is satisfied, so it remains to show that the condition
$s/n= o(1)$ must hold. Note that
\begin{eqnarray*}
 | \!  | \! | {M}_n  | \!  | \!  |_{{2}} & \geq
& \frac{\lambda_n
^2}{n}
 | \!  | \! | (\widehat{\bZ}_S)^T (\widehat{\Sigma
}_{SS})^{-1} \widehat{\bZ}_S  | \!  | \!  |_{{2}}.
\end{eqnarray*}
By definition of the sub-differential of the $\ell_1/\ell_2$ norm, we
have $ | \!  | \! | \widehat{\bZ}_S  | \!  | \!
 |_{{F}}^2 = s$, so that there must
be at least one column of $\widehat{\bZ}_S$ with squared $\ell_2$ norm
greater than $s/K$. Without loss of generality, let us
assume that it is the first column $\widehat{\bZ}_1 \in\mathbb{R}^s$.
We then have
\begin{eqnarray*}
 | \!  | \! | {M}_n  | \!  | \!  |_{{2}} & \geq
& \frac{\lambda_n
^2}{n}
\widehat{\bZ}_1^T (\widehat{\Sigma}_{SS})^{-1} \widehat{\bZ}_1 \\
& \geq& \frac{\lambda_n^2 s}{n K}
\lambda_{\min}( (\widehat{\Sigma}_{SS})^{-1} ) \\
& \geq& \frac{\lambda_n^2 s}{n K}
\frac{1}{\lambda_{\max}(\widehat{\Sigma}_{SS})}.
\end{eqnarray*}
From concentration of random matrix eigenvalues [see
equation~(\ref{EqnMinSpecBound}) in Appendix~\ref{AppRandMat}], we have
$\lambda_{\max}(\widehat{\Sigma}_{SS}) \leq2 \lambda_{\max
}(\Sigma_{S
S})$
with probability greater than $1-\exp(-\Theta(n))$, so that we
conclude that the lower bound $ | \!  | \! | {M}_n  |
\!  | \!  |_{{2}} \geq
\frac{\lambda_n^2 s}{2 K n}$ holds
with high probability (w.h.p.).

Substituting this lower bound into the lower bound~(\ref{EqnVmaxExp})
from Lemma~\ref{LemVmaxExp}, we obtain that w.h.p. for any $\delta'
\in(0,1)$,
%
%
%e52 ###
\begin{eqnarray}
\label{LowerboundExpVmax_slarge}
\frac{1}{\lambda_n} \EE[ V_{\max} | X_S,W] &
\geq
& \sqrt{\frac{ s}{2 K n}} \sqrt{2
(1-\delta')
\rho_{\ell}({\Sigma}_{S^c S^c | S})
\log(p-s)},
\end{eqnarray}
which tends to infinity unless $s/n= o(1)$. By the
concentration around this expected value from Lemma~\ref{LemVmaxLip},
this fact implies that the multivariate group Lasso fails w.h.p. unless
$s/n
= o(1)$.

We have thus shown that the conditions of Lemma~\ref{LemTailBound} are
necessary conditions for the multivariate group Lasso to succeed, and
given that these conditions are satisfied, the quantity
$ | \!  | \! | M_n  | \!  | \!  |_{{2}}$ is
concentrated. Recalling the definition of the
event $\mathcal{T}(\delta)$ from equation~(\ref{EqnDefnTail}), we
can write
\begin{eqnarray*}
\mathbb{P}\biggl[ \frac{V_{\max}}{\lambda_n} \leq2-\mutinc\biggr] &
\leq&
\mathbb{P}[ T'_3 \leq2-\mutinc | \mathcal{T}(\delta)] +
\mathbb{P}[
\mathcal{T}(\delta)^c],
\end{eqnarray*}
where we are guaranteed that $\mathbb{P}[ \mathcal{T}(\delta)^c]
\rightarrow
0$ by Lemma~\ref{LemTailBound}.

Recall that we have established that
$\frac{s}{n}=o(1)$. Conditioned on the event\vspace*{-2pt}
$\mathcal{T}(\delta)$, the inequality $ | \!  | \! |
{M}_n  | \!  | \!  |_{{2}} \geq
\lambda_n^2 \frac{\psi(B^*)}{n}(1-\delta)$ holds;
combined with the lower bound~(\ref{EqnVmaxExp}), for any $\delta'\in
(0,1)$, we have for $(p- s)$ sufficiently large and if
$\frac{s}{n}=o(1)$ that
\begin{eqnarray*}
&&\frac{1}{\lambda_n} \mathbb{E}[V_{\max} |\mathcal{T}(\delta),X_S,W]
\\
&&\qquad \geq \sqrt{\frac{\psi(B^*)}{n} (1-\delta)}
\sqrt{2 (1-\delta') \rho_{\ell}({\Sigma}_{S^c S^c |
S})
\log(p-s)}.
\end{eqnarray*}
Consequently, if the lower bound~(\ref{EqnLowerBound}) holds strictly,
then for $(p- s)$ sufficiently large, denoting $\theta_\ell
:=\rho_{\ell}({\Sigma}_{S^c S^c |
S})/(2-\mutinc)^2$ and $\delta'' :=
\sqrt{(1-\delta')(1-\delta)}-1$ we have
\begin{eqnarray*}
&&\frac{1}{\lambda_n} \mathbb{E}[V_{\max} |\mathcal{T}(\delta),
X_S,
W]
\\
&&\qquad\geq (2-\mutinc) \sqrt{\frac{2 \theta_\ell
\psi(B^*)\log(p-s)}{n}} (1-\delta'')\\
&&\qquad \geq
 \frac{2-\mutinc}{\sqrt{ 1-\nu}} (1-\delta'') \geq(2-\mutinc)
\biggl(1+\frac{\nu}{2}\biggr)(1-\delta'') \geq2-\mutinc+\epsilonn
\end{eqnarray*}
with\footnote{Here we have used the fact that for $\delta,\delta'$
sufficiently small, we have $(1-\delta'')(1+\frac{\nu}{2})\geq
(1+\frac{\nu}{3})$.} $\epsilonn=(2-\mutinc) \frac{\nu}{3}$.

Combining this lower bound with the concentration statement from
Lemma~\ref{LemVmaxLip}, we obtain
\begin{eqnarray*}
\mathbb{P}\biggl[ \frac{V_{\max}}{\lambda_n} \leq2-\mutinc \big|
\mathcal{T}(\delta)\biggr] & \leq& 4 \exp\biggl\{ -\frac{1}{2}
\biggl(\frac{\epsilonn^2}{ \rho_{u }({\Sigma}_{{S^c}{S^c}
\mid
S})} \frac{n}{\psi(B^*) (1-\delta)} \biggr)
\biggr\} \\
& \leq& 4 \exp\biggl\{ -\frac{1}{2} \biggl(\frac{\epsilonn^2 C_{\max}}{
\rho_{u }({\Sigma}_{{S^c}{S^c} \mid S})}
\frac{K
n}{s (1-\delta)} \biggr) \biggr\}\\ & \leq& 4
\exp
\biggl\{- c' \frac{K n}{s} \biggr\},
\end{eqnarray*}
where we have defined the constant $c' :=\frac{(2-\mutinc)^2
C_{\max}}{18\mutinc^2 \theta_u (1-\delta)}$, and used the facts that
$\epsilonn=(2-\mutinc) \frac{\nu}{3}$ and $\theta_u :=
\rho_{u }({\Sigma}_{{S^c}{S^c} \mid
S})/\mutinc^2$. Therefore, the probability vanishes, since the
condition $s/n= o(1)$ is equivalent to $n
/s
\rightarrow+\infty$.

%% BEGIN COMMENT
%% MJW (revision S_l1l2aos.tex)
%% This case is not necessary, since we have shown above that
%% the method will fail unless n/s --> \infty.
%this case, equation~(\ref{LowerboundExpVmax_slarge}) implies that
%%
%X_S,W] \geq(2-\mutinc) ( \sqrt{(1-\delta') \theta_\ell
%
%%
%for some constant $c''> 0$. Consequently, if we set $\epsilonn^2=c''
%%
%).
%%
%Combining the bounds for the cases $\frac{s}{n}=o(1)$ and
%$\frac{n}{s}=\mathcal{O}(1)$, we have thus shown that, under
%the stated conditions in Theorem~\ref{ThmNecessary}, the proposed
%bounds hold.
%}
%%%%%%%%%%%%%%%%%%%%%%%%%%%%%%%%%%%%%%%%%%%%%%%%%%%%%%%%%%%%%%%%%%%%%%%%%%%

%s5 ###
\section{Discussion}
\label{SecDiscuss}

In this paper, we have analyzed the high-dimensional behavior of
block-regularization for multivariate regression problems. Our main
result is to show that that its behavior is governed by the sample
complexity parameter,
\begin{eqnarray*}
\theta_{\ell_1/\ell_2}(n, p, s) & :=& n/[2
\psi(B^*) \log(p- s)],
\end{eqnarray*}
where $n$ is the sample size, $p$ is the ambient dimension
and $\psi(\cdot)$ is a sparsity-overlap function that measures a
combination of the sparsity and overlap properties of the true
regression matrix $B^*$. In particular, Theorems~\ref{ThmMain}
and~\ref{ThmNecessary} show that the multivariate group Lasso either
succeeds (or fails) depending on whether this sample complexity
parameter is larger (or smaller) than a threshold parameter depending
in the design covariance matrix $\Sigma$.

Our results were obtained under high-dimensional scaling, in
particular, assuming the quantities $n, p- s$ and
$s$ all were tending to infinity. As have discussed, the
hypothesis that $s\rightarrow+\infty$ can be relaxed at the
expense of slightly weaker guarantees on the $\ell_2/\ell_\infty$ norm
of the solution. One could also imagine relaxing the constraint
$p-s\rightarrow+\infty$, but for the high-dimensional
problems that motivate our analysis, this is not as interesting, since
in such a case, either the true model is nonsparse (and hence
variable selection is of questionable relevance), or we fall back in
the low-dimensional setting.

There are a number of open questions associated with this work. The
current work applies to the ``hard''-sparsity model, in which a subset
$S$ of the regressors are nonzero, and the remaining coefficients
are zero. As with the ordinary Lasso, it would also be interesting to
study block-regularization under soft sparsity models (e.g., $\ell_q$
``balls'' for coefficients, with $q < 1$). It is also interesting to
consider alternative loss functions such as $\ell_2$ error or
prediction error, as opposed to the exact support recovery criterion
considered here. We note that since this work was first posted, other
researchers have provided related results on consistency in $\ell_2$
error [\citet{lounici2009taking}; \citet{huang2009benefit}], again under hard
sparsity constraints.

%%%%%%%%%%%%%%%%%%%%%%%%%%%%%%%%%%%%%%%%%%%%%%%%%%%%%%%%%%%%%%%%%%%%%%%%%%%%

\begin{appendix}
%%%%%%%%%%%%%%%%%%%%%%%%%%%%%%%%%%%%%%%%%%%%%%%%%%%%%%%%%%%%%%%%%%%%%%%%%%%%%%%

%s6 ###
\section{\texorpdfstring{Proof of
Corollary~2}{Proof of Corollary~2}}
\label{AppIndividual}

Let $\mathcal{F}$ (resp., $\mathcal{F}_0$) be the event that the
thresholded ROLS method fails to recover the
individual supports when applied to the estimated row set $\widehat{S}$
(resp., true row set $S$). By a union bound, the overall
probability of failure in
the multi-stage procedure is upper bounded as $\mathbb{P}[\mathcal{F}]
\leq\mathbb{P}[\widehat{S} \neq S] + \mathbb{P}[\mathcal{F}_0 \mid
\widehat{S} = S]$. Under the conditions of
Theorem~\ref{ThmMain}, the row support is recovered with probability
greater than $1 - \Theta( \exp(-c_0 K\log s))$, so that
$\mathbb{P}[\widehat{S} \neq S] \rightarrow0$. As for the
remaining term, we have\vspace*{-2pt} $\mathbb{P}[\mathcal{F}_0 \mid \widehat{S } =
S] \leq
\frac{\mathbb{P}[\mathcal{F}_0]}{\rule{0pt}{8.5pt}\mathbb{P}[\widehat{S} =S]}$,
which is less than $2 \mathbb{P}[\mathcal{F}_0]$ for $(n,
s)$ large enough, since $\mathbb{P}[\widehat{S} = S]
\rightarrow1$.

Consequently, it suffices to upper bound the unconditional probability
that the ROLS estimate applied to the true support fails to recover
the individual supports. Introducing the shorthand $\widehat
{\Sigma}_{SS}
:=\frac{1}{n}X_{S}^T X_{S}$, some straightforward
linear algebra shows that the ROLS estimate of $B^*_{{S}}$ takes
the form $\widehat{B}_{{S}}=B^*_{{S}}+\widetilde{U}$, where
$\widetilde{U} :=(\widehat{\Sigma}_{SS})^{-\sfrac{1}{2}}
\widetilde{W}/\sqrt{n}$, and $\widetilde{W} :=
(\widehat{\Sigma}_{SS})^{-\sfrac{1}{2}} X_{S}^T W/\sqrt{n}$ is an
$s\times K$ noise matrix with i.i.d. standard Gaussian
entries.

Let $\widetilde{W}^{(j)}$ denote the $j$th column of
$\widetilde{W}$, and let $e_i$ denote the $i$th canonical
basis vector in $\mathbb{R}^{s}$. We then have
\begin{eqnarray*}
\max_{i,j} |\widetilde{U}_{i,j}|=\max_{i,j}
\frac{1}{\sqrt{n}} \big|e_i^T (\widehat{\Sigma}_{SS})^{-\sfrac{1}{2}}
\widetilde{W}^{(j)}\big| & \leq& \frac{1}{\sqrt{n}} \max_{i}
\Bigl[ \| (\widehat{\Sigma}_{SS})^{-\sfrac{1}{2}} e_i\| \max_{j}|\xi
_{i,j}| \Bigr] \\ &
\leq
& \frac{1}{\sqrt{n}} | \! | \! | (\widehat{\Sigma
}_{SS})^{-\sfrac{1}{2}}  | \!  | \!  |_{{2}}
\max_{i,j}|\xi_{i,j}|,
\end{eqnarray*}
where $(\xi_{i,j})$ forms a sequence of identically distributed
standard Gaussian variables (which are dependent in general).
Using a union bound and standard Gaussian tail bounds, for all $\nu> 0$,
we have
\begin{eqnarray*}
\PP\Bigl[\max_{i,j}|\xi_{i,j}| \geq(1+\nu) \sqrt{2 \log(K
s)} \Bigr] & \leq& 2 \exp( - \nu\log(Ks))
\rightarrow 0.
\end{eqnarray*}
A concentration bound for random matrices (see
Appendix~\ref{AppRandMat}) yields
$ | \!  | \! | \widehat{\Sigma}_{SS}^{-\sfrac{1}{2}}
| \!  | \!  |_{{2}}{} \leq\sqrt{2} C_{\min}^{-\sfrac{1}{2}}$ with
probability greater than $1 - \exp(- \Theta(n))$, so that
we obtain
\begin{eqnarray*}
\PP\biggl[\max_{i,j} |\widetilde{U}_{i,j}| \geq(1+\nu) \sqrt{\frac{4
\log Ks}{C_{\min} n}} \biggr] & = &
\mathcal{O}(\exp(-\Theta(\log s))).
\end{eqnarray*}
This result, together with the lower bound on the smallest absolute
value of the nonzero coefficients of $B^*$, shows that the threshold
procedure in step 3 will retain all nonzero coefficients of $B^*$
while correctly setting to zero all entries for which $B^*$ is
actually zero.

%s7 ###
\section{\texorpdfstring{Proof of Lemma~2}{Proof of Lemma~2}}
\label{AppUniqueOpt}

Using the notation $\beta_i$ to denote a row of $B$ and denoting
by
%
%
%e53 ###
\begin{equation}
\mathcal{K} :=\{ (w,v) \in\mathbb{R}^{K} \times\mathbb{R} \mid
\|w\|_2 \leq v \}
\end{equation}
the usual second-order cone (SOC), we can rewrite the original convex
program~(\ref{EqnBlockRegProb}) with $q=2$ as
%
%
%e54 ###
\begin{eqnarray}
\label{EqnBlockRegProb2}
&&\mathop{\min_{B\in\mathbb{R}^{p\times K}}}_{b \in\mathbb{R}^p}  \frac{1}{2 n}
 | \!  | \! | \bY-\bX B  | \!  | \!
 |_{{F}}^2 + \lambda_n\sum_{i=1}^{p
} b_i
\\
&&\qquad\mbox{s.t. }  (\beta_i,b_i) \in\mathcal{K}, 1\leq i \leq
p. \nonumber
\end{eqnarray}
We now dualize the conic constraints~[\citet{Boyd02}], using conic
Lagrange multipliers belonging to the dual cone
$\mathcal{K}^*=\{(\bz,t) \in\mathbb{R}^{K+1}| \bz^T w+vt \geq0,
(w,v) \in\mathcal{K}\}$. The second-order cone $\mathcal{K}$ is
self-dual~[\citet{Boyd02}], so that the convex
program~(\ref{EqnBlockRegProb2}) is equivalent to
\begin{eqnarray*}
&&\mathop{\min_{B\in\mathbb{R}^{p\times K}}}_{ b \in\mathbb{R}^p}
\mathop{\max_{\bZ\in\mathbb{R}^{p\times K}}}_{ t \in\mathbb{R}^p}
 \frac{1}{2 n}
 | \!  | \! | \bY-\bX B  | \!  | \!
 |_{{F}}^2 + \lambda_n\sum_{i=1}^{p
} b_i
- \lambda_n\sum_{i=1}^{p} ( - \bz_i^T \beta_i+
t_i b_i
)\\
&&\qquad\mbox{s.t. }  (\bz_i,t_i) \in\mathcal{K}, 1\leq i
\leq p,
\end{eqnarray*}
where $\bZ$ is the matrix whose $i$th row is $z_i$.

Since the original program is convex and strictly feasible, strong
duality holds and any pair of primal $(B^\star, b^\star)$ and
dual $(\bZ^\star,t^\star)$ solutions has to satisfy the
Karush--Kuhn--Tucker conditions:
\begin{subequation}
%
%
%e59 ###
%e58 ###
%e57 ###
%e56 ###
%e55 ###
\begin{eqnarray}
\|\beta^\star_i\|_2 &\leq& b^\star_i,\qquad 1<i<p,\label{PrimConstraints}\\
\|\bz^\star_i\|_2 &\leq& t^\star_i,\qquad 1<i<p,\label{DualConstraints}\\
{\bz^\star_i}^T \beta^\star_i-t^\star_i b^\star_i&=&0,\qquad 1<i<p,\label{CompSlack}\\
\nabla_{B}  \biggl[\frac{1}{2 n} | \!  | \! | \bY-\bX B  | \!  | \! |_{{F}}^2 \biggr]\bigg |_{B=B^\star} +
\lambda_n\bZ^\star&=&0, \label{GradPrimal}\\
\lambda_n(1-t^\star_i)&=&0 .\label{ti1}
\end{eqnarray}
\end{subequation}
Since equations~(\ref{CompSlack}) and~(\ref{ti1}) impose the
constraints $t^\star_i=1$ and $b^\star_i=\|\beta^\star_i\|_2$, a~primal--dual solution to this conic program is determined by
$(B^\star,\bZ^\star)$.

Any solution satisfying the conditions in Lemma~\ref{LemUniqueOpt}
also satisfies these KKT conditions, since
equation~(\ref{EqnOnPrimal}) and the definition~(\ref{EqnOffDual}) are
equivalent to equation~(\ref{GradPrimal}), and
equation~(\ref{EqnOnDual}) and the combination of
conditions~(\ref{EqnOffPrimal}) and~(\ref{EqnOffDual}) imply that the
complementary slackness equations~(\ref{CompSlack}) hold for each
primal--dual conic pair $(\beta_i,\bz_i)$.

Now consider some other primal solution $\widetilde{B}$; when combined with
the optimal dual solution $\widehat{\bZ}$, the pair $(\widetilde
{B},\widehat{\bZ})$
must satisfy the KKT conditions~[\citet{Bertsekasnonlin}]. But since
for $j \in{S^c}$, we have $\|\hat{\bz}_j\|_2 < 1$, then the
complementary slackness condition~(\ref{CompSlack}) implies that for
all $j \in{S^c}, \widetilde{\beta}_j=0$. This fact in turn
implies that the primal solution $\widetilde{B}$ must also be a
solution to
the restricted convex program~(\ref{EqnRestricted}), obtained by only
considering the covariates in the set $S$ or equivalently by
setting $B_{S^c}= 0_{S^c}$. But since $s
<n$
by assumption, the matrix $\bX _S^T \bX _S$ is strictly positive
definite with probability one, and therefore the restricted convex
program~(\ref{EqnRestricted}) has a unique solution
$B^\star_S=\widehat{B}_S$. We have thus shown that a
solution $(\widehat{B},\widehat{\bZ})$ to the program~(\ref
{EqnBlockRegProb})
that satisfies the conditions of Lemma~\ref{LemUniqueOpt}, if it
exists, must be unique.

%%%%%%%%%%%%%%%%%%%%%%%%%%%%%%%%%%%%%%%%%%%%%%%%%%%%%%%%%%%%%%%%%%%%%%%%%%%
%s8 ###
\section{Characterization of the sparsity-overlap function}
\label{AppLemSparseOverlap}

In this appendix, we prove Lemma~\ref{LemSparseOverlap}.
(a) To verify this claim, we first set ${\bZ^*_S}= \zeta(B^*_S)$,
and use $Z_S^{(k)*}$ to denote the $k$th column of
${\bZ^*_S}$. Since the spectral norm is upper bounded by the sum of
eigenvalues, and lower bounded by the average eigenvalue, we have
\[
\frac{1}{K} \operatorname{tr}({{\bZ^*_S}}^T\Sigma_{SS}^{-1}{{\bZ^*_S}})
\leq\psi(B^*) \leq\operatorname{tr}({{\bZ^*_S}}^T\Sigma_{S
S}^{-1}{{\bZ^*_S}}).
\]
Given our assumption (A1) on $\Sigma_{SS}$, we have
\[
\operatorname{tr}({{\bZ^*_S}}^T\Sigma_{SS}^{-1}{{\bZ^*_S}})=
\sum_{k=1}^{K} {Z_S^{(k)*}}^T \Sigma_{SS}^{-1}
Z_S^{(k)*} \geq\frac{1}{C_{\max}} \sum_{k=1}^{K}
\big\|Z_S^{(k)*}\big\|^2 = \frac{s}{C_{\max}},
\]
using the fact that $\sum_{k=1}^{K} \|Z_S^{(k)*}\|^2 =
\sum_{i=1}^s\|\bZ^*_i\|^2 = s$. Similarly, in the
other direction, we have
\[
\operatorname{tr}({{\bZ^*_S}}^T\Sigma_{SS}^{-1}{{\bZ^*_S}})=
\sum_{k=1}^{K} {Z_S^{(k)*}}^T \Sigma_{SS}^{-1}
Z_S^{(k)*} \leq\frac{1}{C_{\min}} \sum_{k=1}^{K}
\big\|Z_S^{(k)*}\big\|^2 = \frac{s}{C_{\min}},
\]
which completes the proof.

(b) Under the assumed orthogonality, the matrix ${\bZ^*}^T\bZ^*$ is
diagonal with $\|Z^{(k)*}\|^2$ as the diagonal elements, so that the
largest $\|Z^{(k)*}\|^2$ is then the largest eigenvalue of the
matrix.

%%%%%%%%%%%%%%%%%%%%%%%%%%%%%%%%%%%%%%%%%%%%%%%%%%%%%%%%%%%%%%%%%%%%%%%%%%%

%s9 ###
\section{Group Lasso versus ordinary Lasso}
\label{AppCorGroupOrd}

In this appendix, we provide the proof of Corollary~\ref{CorGroupOrd}
which characterizes the relative efficiency of the group versus the
ordinary Lasso. From the discussion preceding the statement of
Corollary~\ref{CorGroupOrd}, we know that the quantity
\[
\max_{k=1, \ldots, K} \psi\bigl(\beta_S^{*(k)}\bigr) \log
(p- s_k) = \max_{k=1, \ldots, K} s_k \log
(p- s_k) \geq \max_{k=1, \ldots, K} s_k
\log
(p- s)
\]
governs the performance of the ordinary Lasso procedure for row
selection. It remains to show then that $\psi(B^*_S) \leq\max_{k}
s_k$.

As before, we use the notation ${\bZ^*_S}=\zeta(B^*_S)$, and $Z^*_i$
for the $i$th row of ${\bZ^*_S}$. Since $\Sigma_{S
S} = I_{s\times s}$, we have $\psi(B^*) =
\| {\bZ^*_S}\|^2$. Consequently, by the variational representation of
the $\ell_2$-norm, we have
\[
\psi(B^*) = \max_{x \in\mathbb{R}^{K} \dvtx  \|x\|
\leq1}
\| {\bZ^*_S} x\|^2 \leq\max_{x \in\mathbb{R}^{K} \dvtx  \|x\| \leq1}
\sum_{i=1}^s ({Z^*_i}^T x )^2.
\]
Let $|Z^*_i|=(|Z^*_{i1}|,\ldots,|Z^*_{ik}|)^T$ and $y_i=
(x_1 \operatorname{sign}(Z^*_{i1}),\ldots,x_{K} \operatorname
{sign}(Z^*_{i K})
)^T$. By the Cauchy--Schwarz inequality,
\[
({Z^*_i}^T x )^2= ({|Z^*_i|}^T y_i )^2\leq
\| |Z^*_i| \|^2 \|y_i\|^2=\|Z^*_i\|^2 \sum_k x_k^2
\operatorname{sign}(Z^*_{ik})^2
\]
so that
\[
\sum_{i=1}^s ({Z^*_i}^T x )^2 \leq\sum_{i=1}^s
\|Z^*_i\|^2 \sum_{k=1}^Kx_k^2
\operatorname{sign}(Z^*_{ik})^2=\sum_{k=1}^Kx_k^2 \sum_{i=1}^s
\operatorname{sign}(Z^*_{ik})^2=\sum_{k=1}^Kx_k^2 s_k,
\]
and if $\|x\| \leq1$, we have $\sum_{k=1}^Kx_k^2 s_k \leq
\max_{1 \leq k \leq K} s_k$ thereby establishing the claim.

%%%%%%%%%%%%%%%%%%%%%%%%%%%%%%%%%%%%%%%%%%%%%%%%%%%%%%%%%%%%%%%%%%%%%%%%

%s10 ###
\section{Inequalities with block-matrix norms}
\label{AppLemGuillaume}
In general, the two families of matrix norms that we have introduced,
$|\!|\!| \cdot | \! | \!|_{{p}, {q}}$ and
$\| \cdot \|_{\ell_{a}/\ell_{b}}$, are distinct,
but they coincide in the following useful special case:
\begin{lems}
\label{LemGuillaume}
For $1 \leq p \leq\infty$ and for $r$ defined by $1/r+1/p=1$ we have
\begin{eqnarray*}
\| \cdot \|_{\ell_{\infty} /\ell_{p}} & = & |\!|\!| \cdot | \! | \!
|_{{\infty}, {r}}.
\end{eqnarray*}
\end{lems}

\begin{pf}
Indeed, if $a_i$ denotes the $i$th row of $A$, then
\begin{eqnarray*}
\| A \|_{\ell_{\infty} /\ell_{p}}&=&\max_i \|a_i\|_p=\max_i \max_{\|
y_i\|_r
\leq
1} y_i^T a_i
\\
&=&\max_{\|y\|_r \leq1} \max_i |y^T a_i|=\max_{\|y\|_r
\leq
1} \|Ay\|_\infty.
%=\specmatnorm{A}{\infty}{r}.
\end{eqnarray*}\upqed
\end{pf}

We conclude by stating some useful bounds and relations:
\begin{lems}
\label{LemGuillaume1}
Consider matrices $A \in\mathbb{R}^{m \times n}$ and $\bZ\in\mathbb{R}^{n
\times
\ell}$ and $p, r > 0$ with $\frac{1}{p}+\frac{1}{r}=1$, we have
\begin{subequation}
%
%
%e61 ###
%e60 ###
\begin{eqnarray}
\label{EqnSubordinate}
\hspace*{20pt}\| A \bZ \|_{\ell_{\infty} /\ell_{p}} & = & |\!|\!| A \bZ | \! | \!
|_{{\infty}, {r}} \leq
|\!|\!| A | \! | \!|_{{\infty}, {\infty}} |\!|\!| Z | \! | \!
|_{{\infty}, {r}} =
|\!|\!| A | \! | \!|_{{\infty}, {\infty}} \| \bZ \|_{\ell_{\infty} /\ell_{p}}, \\
\label{EqnGuillaume2}
 | \! | \! | A  | \!  | \!  |_{{r}} & \leq&|\!
|\!| I_m | \! | \!|_{{r}, {\infty}}
|\!|\!| A | \! | \!|_{{\infty}, {r}} =s^{1/r} \| A \|_{\ell_{\infty} /\ell_{p}}.
\end{eqnarray}
\end{subequation}
\end{lems}
%

%%%%%%%%%%%%%%%%%%%%%%%%%%%%%%%%%%%%%%%%%%%%%%%%%%%%%%%%%%%%%%%%%%%%%%%
%s11 ###
\section{Some concentration inequalities for random matrices}
\label{AppRandMat}

In this appendix, we state some known concentration inequalities for
the extreme eigenvalues of Gaussian random matrices.
Although these results hold more generally, our interest here is on
scalings $(n, s)$ such that $s/n
\rightarrow
0$. The following result is from~\citet{DavSza01}.
%l9
\begin{lems}
Let $U \in\mathbb{R}^{n\times s}$ be a random matrix from
the standard Gaussian ensemble [i.e., $U_{ij} \sim N(0,1)$, i.i.d.].
Then if we denote by $\lambda_{\min}(\cdot)$ and $\lambda_{\max
}(\cdot)$
the smallest and largest singular value of $U$, respectively, we have
%
%
%e63 ###
%e62 ###
\begin{eqnarray}
\mathbb{P}\Biggl[1-\lambda_{\min} \biggl(\frac{U}{\sqrt{n}} \biggr)
\geq\sqrt{\frac{s}{n}}+t \Biggr] \leq\exp\biggl(-
\frac{nt^2}{2} \biggr),\\
\mathbb{P}\Biggl[ \lambda_{\max}
\biggl(\frac{U}{\sqrt{n}} \biggr)-1\geq\sqrt{\frac{s
}{n}}+t \Biggr] \leq\exp\biggl(- \frac{nt^2}{2} \biggr).
\end{eqnarray}
\end{lems}

As a consequence, for $s/n\rightarrow0$ we obtain the
two following inequalities:
\begin{lems}
%
%
%e65 ###
%e64 ###
\begin{eqnarray}
\label{EqnMinSpecBound}
\mathbb{P}\biggl[ \bigg| \!  \bigg| \! \bigg| \frac{1}{n} U^T U  \bigg| \! \bigg| \!  \bigg|_{{2}}{} \leq\frac
{1}{2} \biggr] & \leq& \exp\Biggl\{ -\frac{n}{2} \Biggl(\frac
{1}{4}-\sqrt{\frac{s}{n}} \Biggr)_+^2 \Biggr\} ,\\
\label{EqnRandBound}
\mathbb{P}\Biggl[ \bigg| \!  \bigg| \! \bigg| \frac{1}{n} U^T U - I_{s\times
s}  \bigg| \!  \bigg| \!  \bigg|_{{2}}{} \geq6 \sqrt{\frac{s}{n}} \Biggr] &
\leq& 2
\exp\biggl(- \frac{s}{2} \biggr) + \exp(-\Theta(n))
\rightarrow 0.
\end{eqnarray}
\end{lems}
\begin{pf}
For simplicity, we write $\lambda_{\min}$ for $\lambda_{\min}
(\frac{U}{\sqrt{n}} )$ and $\lambda_{\max}$ for $\lambda
_{\max} (\frac{U}{\sqrt{n}} )$.
For equation (\ref{EqnMinSpecBound}), we have
\begin{eqnarray*}
\mathbb{P}\biggl[ \bigg| \!  \bigg| \! \bigg| \frac{1}{n} U^T U  \bigg| \! \bigg| \!  \bigg|_{{2}}{}
\leq\frac{1}{2} \biggr]
&\leq&\mathbb{P}\biggl[ \lambda_{\min}
\leq\frac{1}{\sqrt{2}} \biggr] \leq\mathbb{P}\Biggl[ 1 - \lambda_{\min} \geq\sqrt{\frac{s}{n}}+ \Biggl(\frac{1}{4}-\sqrt{\frac{s}{n}} \Biggr) \Biggr]
\\
&\leq&\exp\Biggl\{ -\frac{n}{2} \Biggl(\frac{1}{4}-\sqrt{\frac{s}{n}} \Biggr)_+^2 \Biggr\}.
\end{eqnarray*}
For equation~(\ref{EqnRandBound}),
\begin{eqnarray*}
& & \mathbb{P}\Biggl[ \bigg| \!  \bigg| \! \bigg| \frac{1}{n} U^T U - I_{s
\times s}  \bigg| \!  \bigg| \!  \bigg|_{{2}}{} \geq6 \sqrt{\frac{s}{n}}
\Biggr]
\\
&&\qquad=
\mathbb{P}\Biggl[\max ( \lambda_{\max}^2-1 ,1-\lambda_{\min}^2
) \geq6 \sqrt{\frac{s}{n}} \Biggr]\\
&&\qquad\leq \mathbb{P}\Biggl[ \lambda_{\max} -1 \geq2 \sqrt{\frac
{s}{n}} \Biggr] +
\mathbb{P}\Biggl[ 1- \lambda_{\min} \geq2 \sqrt{\frac{s
}{n}} \Biggr]
\\
&&\qquad\quad{}+ \mathbb{P}[ \lambda_{\max} + 1 \geq3 ] +
\mathbb{P}[ \lambda_{\min}+1 \geq3 ]
\\
&&\qquad\leq 2 \exp \Biggl\{ -\frac{n}{2} \Biggl( \sqrt{\frac
{s}{n}} \Biggr)^2 \Biggr\} +2 \exp\Biggl\{ -\frac
{n}{2} \Biggl( \frac{1}{4}-\sqrt{\frac{s}{n}}
\Biggr)_+^2 \Biggr\},
\end{eqnarray*}
where we used that $ \{ \lambda^2-1 \geq x \} \subset\{\lambda-1
\geq\frac{x}{3} \} \cup\{\lambda+1 \geq3 \}$ to obtain the first
inequality.
\end{pf}

These results are easily adapted to more general Gaussian ensembles.
Letting $X = U \sqrt{\Lambda}$, we obtain an $n\times s$
matrix with i.i.d.\ rows, $X_i \sim N(0, \Lambda)$. If the covariance
matrix $\Lambda$ has maximum eigenvalue $C_{\max}< +\infty$, then we
have
%
%
%e66 ###
\begin{eqnarray}
\label{EqnGenGaussRandBound}
\hspace*{30pt} | \!  | \! | n^{-1} X^T X - \Lambda  | \!  | \! |_{{2}}{} & = &
 \big| \!  \big| \! \big| \sqrt{\Lambda} [n^{-1} U^T U - I ] \sqrt
{\Lambda}  \big| \!  \big| \!  \big|_{{2}}{} \leq C_{\max} | \!
 | \! | n^{-1} U^T U - I  | \!  | \!  |_{{2}}{}
\end{eqnarray}
so that the bound~(\ref{EqnRandBound}) immediately yields an analogous
bound on different constants.

The final type of bound that we require is on the difference
\begin{eqnarray*}
 | \!  | \! | (X^T X/n)^{-1} - \Lambda^{-1}  | \!
| \!  |_{{2}}{},
\end{eqnarray*}
assuming that $X^T X$ is invertible. We note that
\begin{eqnarray*}
 | \!  | \! | (X^T X/n)^{-1} - \Lambda^{-1}  | \!
| \!  |_{{2}}{} & = &  | \!  | \! | (X^T X/n)^{-1}
[\Lambda- (X^T X/n)] \Lambda^{-1}  | \!  | \!  |_{{2}}{}
\\
& \leq&  | \!  | \! | (X^T X/n)^{-1}  | \!  | \!
 |_{{2}}{} | \!  | \! | \Lambda- (X^T X/n)  | \!
 | \!  |_{{2}}{}  | \!  | \! | \Lambda^{-1}  |
\!  | \!  |_{{2}}{}.
\end{eqnarray*}
As long as the eigenvalues of $\Lambda$ are bounded below by
$C_{\min}> 0$, then\break\mbox{$ | \!  | \! | \Lambda^{-1}  | \!  | \!
 |_{{2}}{} \leq1/C_{\min}$}. Moreover, since
$s/n\rightarrow0$, we have [from
equation~(\ref{EqnMinSpecBound})] that $ | \!  | \! | (X^T
X/ n)^{-1}  | \!  | \!  |_{{2}}{} \leq2/C_{\min}$ with
probability converging to one
exponentially in $n$. Thus,
equation~(\ref{EqnGenGaussRandBound}) implies the desired bound.
%%%%%%%%%%%%%%%%%%%%%%%%%%%%%%%%%%%%%%%%%%%%%%%%%%%%%%%%%%%%%%%%%%%%%%%%

%s12 ###
\section{\texorpdfstring{Proof of Lemma~3}{Proof of Lemma~3}}
\label{AppLemZapprox}

The analysis in Section~\ref{SecUevent} shows that the condition
$\|{\Delta}_i\|_2 \leq1/2$ implies that $\widehat{\beta}_i \neq
\vec{0}$ and
hence $\widehat{\bZ}_i = \widehat{\beta}_i/\|\widehat{\beta
}_i\|_2$ for all rows
$i \in S$. Therefore, using the notation
$Z^*_i=\beta^*_i/\|\beta^*_i\|_2$ we have
\begin{eqnarray*}
\widehat{\bZ}_i - Z^*_i & = &
\frac{\widehat{\beta}_i}{\|\widehat{\beta}_i\|_2}-Z^*_i
=\frac{Z^*_i+ {\Delta}_i} {\|Z^*_i+
{\Delta}_i\|_2}-Z^*_i\\
& = & Z^*_i \biggl( \frac{1}
{\|Z^*_i+ {\Delta}_i\|_2} - 1 \biggr)+\frac{ {\Delta}_i}
{\|Z^*_i+ {\Delta}_i\|_2}.
\end{eqnarray*}
Note that, for $z \neq0$, the function
$g(\bz,{\delta})=\frac{1}{\| \bz+ \ensuremath
{{\delta}}\|_2}$ is differentiable
with respect to~${\delta}$, with gradient $\nabla
_{ {\delta}}
g(\bz,{\delta})=-\frac{\bz+\ensuremath{{\delta
}}}{2\|\bz+{\delta}\|_2^3}$. By
the mean-value theorem, there exists $h \in[0,1]$ such that
\[
\frac{1}{\| \bz+ {\delta}\|_2}-1=g(\bz
,{\delta})-g(\bz,0)=\nabla_{ \ensuremath
{{\delta}}}
g(\bz,h{\delta})^T \ensuremath{{\delta
}}=-\frac{(\bz+h {\delta})^T
{\delta}}{2\|\bz+h {\delta}\|_2^3},
\]
which implies that there exists $h_i\in[0,1]$ such that
%
%
%e67 ###
\begin{eqnarray}\label{EqnHelpful}
\|\widehat{\bZ}_i - Z^*_i\|_2 & \leq& \| Z^*_i\|_2 \frac{|(Z^*_i+h_i
{\Delta}_i)^T {\Delta}_i|}{2\|Z^*_i+h_i {\Delta}_i\|_2^3}+\frac{ \|
{\Delta}_i\|_2} {\|Z^*_i+ {\Delta}_i\|_2} \nonumber\\[-8pt]\\[-8pt]
& \leq&
\frac{\| {\Delta}_i\|_2}{2\|Z^*_i+h_i {\Delta}_i\|_2^2}+\frac{ \|
{\Delta}_i\|_2} {\|Z^*_i+ {\Delta}_i\|_2}.\nonumber
\end{eqnarray}
We note that $\| Z^*_i\|_2=1$ and $\|{\Delta}_i\|_2 \leq
\frac{1}{2}$ imply that $\|Z^*_i+h_i {\Delta}_i\|_2 \geq
\frac{1}{2}$. Combined with inequality~(\ref{EqnHelpful}), we obtain
$\|\widehat{\bZ}_i - Z^*_i\|_2 \leq4 \|{\Delta}_i\|_2$, which
proves the lemma.

%%%%%%%%%%%%%%%%%%%%%%%%%%%%%%%%%%%%%%%%%%%%%%%%%%%%%%%%%%%%%%%%%%%%%%%%%%%%%%%

%s13 ###
\section{\texorpdfstring{Proof of Lemma~4}{Proof of Lemma~4}}
\label{AppLemTailBound}

With $\bZ^*_S=\zeta(B^*_S)$, define the $K
\times
K$ random matrix
\begin{eqnarray*}
{M}^*_n& :=& \frac{\lambda_n^2}{n}
(\bZ^*_S)^T (\widehat{\Sigma}_{SS})^{-1} \bZ^*_S+
\frac{1}{n^2} W^T (I_n- \Pi_{S}) W
\end{eqnarray*}
and note that (using standard results on Wishart
matrices~[\citet{AndersonStat}])
%
%
%e68 ###
\begin{eqnarray}
\label{EqnWishMean}
\EE[{M}^*_n] &=& \frac{\lambda_n^2}{n
-s-1}
(\bZ^*_S)^T (\Sigma_{SS})^{-1} \bZ^*_S+
\sigma^2
\frac{n-s}{n^2} I_K.
\end{eqnarray}

To bound ${M}_n$ in spectral norm, we use the triangle
inequality,
%
%
%e69 ###
\begin{eqnarray}
\label{EqnBigTriangle}
\big|  | \!  | \! | {M}_n  | \!  | \!  |_{{2}}{}
- | \!  | \! | \EE[{M}^*_n]  | \!  | \!
|_{{2}}{} \big| & \leq&
 | \!  | \! | {M}_n-\EE[{M}^*_n]  | \!  | \!
|_{{2}}{} \nonumber\\[-8pt]\\[-8pt]
& \leq & \mathop{\underbrace{{ | \!  | \! | {M}_n- {M}^*_n  |\!  | \!  |_{2}}_{}}}_{A_1}
+ \mathop{\underbrace{{ | \!  | \! |{M}^*_n- \EE[{M}^*_n]  | \!  | \!  |_{2}}_{}}}_{A_2}.\nonumber
%
%& & A_1 A_2 \nonumber
\end{eqnarray}
Considering the term $A_1$ in the
decomposition~(\ref{EqnBigTriangle}), we have
%
%
%e70 ###
\begin{eqnarray}\label{EqnAoneBound}
\hspace*{50pt}&& | \! | \! | {M}_n^*-{M}_n  | \!  | \!  |_{{2}}\nonumber
\\
&&\qquad =
\frac{\lambda_n^2}{n}  | \! | \! | \bZ^*_S\widehat
{\Sigma}_{SS}^{-1} \bZ^*_S-\widehat{\bZ}_S\widehat{\Sigma
}_{SS}^{-1} \widehat{\bZ}_S  | \!  | \!  |_{{2}}
\nonumber\\[-8pt]\\[-8pt]
&&\qquad =  \frac{\lambda_n^2}{n}  \big| \! \big| \! \big| \bZ^*_S
\widehat{\Sigma}_{SS}^{-1} (\bZ^*_S-\widehat{\bZ}_S) +(\bZ^*_S
-\widehat{\bZ}_S) \widehat{\Sigma}_{SS}^{-1} \bigl(\bZ^*_S +(\widehat
{\bZ}_S-\bZ^*_S)\bigr)  \big| \!  \big| \!  \big|_{{2}} \nonumber\\
&&\qquad \leq \frac{\lambda_n^2}{n}  | \! | \! | \widehat
{\Sigma}_{SS}^{-1}  | \!  | \!  |_{{2}}
 | \! | \! | \bZ^*_S-\widehat{\bZ}_S  | \!  | \!
 |_{{2}} (2  | \! | \! | \bZ^*_S  | \!  | \!
 |_{{2}}+ | \! | \! | \bZ^*_S -\widehat{\bZ}_S
| \!  | \!  |_{{2}} ).\nonumber
\end{eqnarray}
Using the concentration results on random matrices in
Appendix~\ref{AppRandMat}, we have the bound
$ | \! | \! | \widehat{\Sigma}_{SS}^{-1}  | \!
 | \!  |_{{2}} \leq2/C_{\min}$ with probability
greater than $1-\exp(- \Theta(n))$, and we have
$ | \! | \! | \bZ^*_S  | \!  | \!
|_{{2}}=\mathcal{O}(\sqrt{s})$ by
definition. Moreover, from equation~(\ref{EqnGuillaume2}) in
Lemma~\ref{LemGuillaume}, we have $ | \! | \! | \bZ
^*_S-\widehat{\bZ}_S  | \!  | \!  |_{{2}} \leq\sqrt{s}
\| \bZ^*_S-\widehat{\bZ}_S \|_{\ell_{\infty} /\ell_{2}}$. Using the
bound~(\ref{EqnKeyUBound}) and Lemma~\ref{LemZapprox}, we have
$\| \bZ^*_S-\widehat{\bZ}_S \|_{\ell_{\infty} /\ell_{2}} = o(1)$ with
probability greater than $1-c_1 \exp(-c_0 K\log s)$, so
that from
equation~(\ref{EqnAoneBound}), we conclude that
%
%
%e71 ###
\begin{eqnarray}
\label{EqnAtermOne}
A_1 & = &  | \! | \! | {M}_n^*-{M}_n  | \!  | \!
 |_{{2}} =
o \biggl(\frac{\lambda_n^2 s}{n} \biggr)\qquad
\mbox{w.h.p.}
\end{eqnarray}

Turning to term $A_2$, we have the upper bound $A_2
\leq T^\dagger_1 + T^\dagger_2$, where
\begin{eqnarray*}
T^\dagger_1 & :=& \frac{\lambda_n^2}{n}
 | \! | \! | \bZ^*_S  | \!  | \!  |_{{2}}^2
 \bigg| \! \bigg| \! \bigg| \frac{n }{n- s-1} (\Sigma_{SS})^{-1} -
(\widehat{\Sigma}_{SS })^{-1} \bigg | \!  \bigg| \!  \bigg|_{{2}}\quad
\end{eqnarray*}
and
\begin{eqnarray*}
T^\dagger_2 & :=& \frac{1}{n^2}  | \! | \! | W^T (I_n-
\Pi_{S}) W- \sigma^2 (n- s) I_K  | \!  | \!  |_{{2}}.
\end{eqnarray*}
Since\vspace*{-2pt} $ | \! | \! | \bZ^*_S  | \!  | \!
|_{{2}}^2 \leq s$, and $ | \! | \! | \frac{n}{n- s-1}
(\Sigma_{SS})^{-1} - (\widehat{\Sigma}_{SS})^{-1}  | \!  |
\!  |_{{2}} = o(1)$ with high probability
(see Appendix~\ref{AppRandMat}), we
have $T^\dagger_1 = o ( \frac{\lambda_n^2 s}{n}
)$ with probability greater than $1-2 \exp(-\Theta(n))$.

Turning to $T^\dagger_2$, we have with probability greater than
$1-2\exp(- s/2)-\exp(- \Theta(n))$,
\begin{eqnarray*}
T^\dagger_2 & = & \mathcal{O} \biggl( \frac{\sqrt{s}}{ n\sqrt
{n}} \biggr) = o \biggl( \frac{1}{n} \biggr),
\end{eqnarray*}
using the random matrix bound~(\ref{EqnRandBound}) once again.
Overall, we conclude that
%
%
%e72 ###
\begin{eqnarray}
\label{EqnAtermTwo}
A_2 =  | \! | \! | {M}^*_n-\EE[{M}^*_n ]  | \!  |
\!  |_{{2}} &
= & o \biggl( \frac{\lambda^2_n s+ 1}{n} \biggr)\qquad
\mbox{w.h.p.}
\end{eqnarray}

Finally, turning to $ | \!  | \! | \EE[{M}^*_n]  | \!
 | \!  |_{{2}}{}$,
from equation~(\ref{EqnWishMean}), we have
%
%
%e73 ###
\begin{eqnarray}
\label{EqnMstarExp}
 | \!  | \! | \mathbb{E}[{M}^*_n]  | \!  | \!
|_{{2}}{} &=&
\frac{\lambda_n^2 \psi(B^*)}{n}
\frac{n}{n-s-1} + \frac{\sigma^2}{n}
\biggl(1-\frac{s}{n} \biggr)\nonumber
\\[-8pt]\\[-8pt]
&=& \bigl(1 + o(1) \bigr)
\biggl[\frac{\lambda^2_n \psi(B^*) + \sigma
^2}{n} \biggr].\nonumber
\end{eqnarray}

Finally, we combine bounds~(\ref{EqnAtermOne}), (\ref{EqnAtermTwo})
and~(\ref{EqnMstarExp}) in the decomposition~(\ref{EqnBigTriangle}),
and apply Lemma~\ref{LemSparseOverlap}(a) to obtain that
$\psi(B^*) = \Theta(s)$; combining these facts
yields that
\begin{eqnarray*}
(1 - \delta) \biggl[\frac{\lambda^2_n
\psi(B^*)+\sigma^2}{n} \biggr] \leq
 | \!  | \! | {M}_n  | \!  | \!  |_{{2}}{} &
\leq& (1 + \delta)
\biggl[\frac{\lambda^2_n \psi(B^*)+\sigma
^2}{n}
\biggr]
\end{eqnarray*}
with probability greater than $1 - c_1 \exp(-c_0 K\log s
)$, which
establishes the claim.

%s14 ###
\section{Large deviations for $\chi^2$-variates}
\label{AppChiMax}

\begin{lems}
\label{LemChiMax}
Let $Z_1, \ldots, Z_m$ be i.i.d.\ $\chi^2$-variates with $d$ degrees
of freedom. Then for all $t > d$, we have
%
%
%e74 ###
\begin{eqnarray}
\label{EqnChiMax}
\mathbb{P}\Bigl[\max_{i=1, \ldots, m} Z_i \geq2t\Bigr] & \leq& m \exp\Biggl(-t
\Biggl[1 - 2 \sqrt{\frac{d}{t}} \Biggr] \Biggr).
\end{eqnarray}
\end{lems}
\begin{pf}
Given a central $\chi^2$-variate $X$ with $d$ degrees of
freedom, \citet{LauMas98} prove that $\mathbb{P}[X -
d\geq2 \sqrt{d x} + 2x] \leq\exp(-x)$, or equivalently
\begin{eqnarray*}
\mathbb{P}\bigl[X \geq x + \bigl(\sqrt{x} + \sqrt{d}\bigr)^2 \bigr] & \leq&
\exp(-x),
\end{eqnarray*}
valid for all $x > 0$. Setting $\sqrt{x} + \sqrt{d} = \sqrt{t}$,
we have
\begin{eqnarray*}
\mathbb{P}[X \geq2 t] &\stackrel{\mathrm{(a)}}{\leq}& \mathbb{P}\bigl[X \geq
\bigl(\sqrt{t}- \sqrt{d}\bigr)^2 + t \bigr]
 \leq \exp\bigl(-\bigl(\sqrt{t} -\sqrt{d}\bigr)^2\bigr) \\
& \leq& \exp\bigl(-t + 2 \sqrt{t d}\bigr) \\
& = & \exp\Biggl(-t \Biggl[1- 2 \sqrt{\frac{d}{t}} \Biggr] \Biggr),
\end{eqnarray*}
where inequality (a) follows since $\sqrt{t} \geq\sqrt{d}$ by
assumption. Thus, the claim~(\ref{EqnChiMax}) follows by the union
bound.
\end{pf}
\end{appendix}

\printaddresses

\end{document}